\documentclass[]{fairmeta}

\usepackage[utf8]{inputenc} %
\usepackage[T1]{fontenc}    %
\usepackage{url}            %
\usepackage{booktabs}       %
\usepackage{amsfonts}       %
\usepackage{nicefrac}       %
\usepackage{microtype}      %
\definecolor{bluelink}{RGB}{0,113,188}
\definecolor{greenlink}{RGB}{0,188,113}
\definecolor{PineGreen}{RGB}{0.0, 0.47, 0.44}
\definecolor{Gray}{RGB}{0.5,0.5,0.5}
\usepackage{listings} 
\usepackage{wrapfig}
\usepackage[most]{tcolorbox}

\newtcolorbox{samplebox}[1]{
    breakable, 
    colback=blue!5!white,
    colframe=blue!50!black, 
    fonttitle=\bfseries,
    title=#1
}
\definecolor{citecolor}{HTML}{0071bc}
\hypersetup{
    colorlinks=true,%
    citecolor=citecolor,%
    filecolor=citecolor,%
    linkcolor=citecolor,%
    urlcolor=citecolor
}
\usepackage{tabularx}
\usepackage[most]{tcolorbox}
\usepackage{amsmath}
\usepackage{multirow}
\usepackage{array}
\usepackage{caption}
\usepackage{wrapfig}
\usepackage{enumitem}
\usepackage{tikz}
\usepackage{lipsum}
\usepackage{subcaption}
\usepackage{multirow} 
\usepackage{adjustbox}
\usepackage[table]{xcolor}  
\usepackage{siunitx}          
\usepackage{graphicx}     
\newcommand\eg{\emph{e.g.}} 
\newcommand\ie{\emph{i.e.}}

\usepackage{amssymb}
\renewcommand{\paragraph}[1]{\vspace{1.25mm}\noindent\textbf{#1}}

\usepackage{colortbl}
\definecolor{lightyellow}{rgb}{1.0, 0.98, 0.8}
\definecolor{lightblue}{rgb}{0.85, 0.9, 1.0}
\definecolor{lightgreen}{rgb}{0.85, 1.0, 0.9}
\definecolor{nvidiagreen}{rgb}{0.4627, 0.7255, 0.0}
\definecolor{lightorange}{rgb}{0.99, 0.83, 0.64}
\definecolor{baselinecolor}{rgb}{0.9, 0.9, 0.9}
\usepackage{soul}
\sethlcolor{nvidiagreen}

\newlength\savewidth\newcommand\shline{\noalign{\global\savewidth\arrayrulewidth
  \global\arrayrulewidth 1pt}\hline\noalign{\global\arrayrulewidth\savewidth}}
\newcolumntype{x}[1]{>{\centering\arraybackslash}p{#1pt}}
\newcolumntype{y}[1]{>{\raggedright\arraybackslash}p{#1pt}}
\newcolumntype{z}[1]{>{\raggedleft\arraybackslash}p{#1pt}}
\newcommand{\tablestyle}[2]{\setlength{\tabcolsep}{#1}\renewcommand{\arraystretch}{#2}\centering\footnotesize}

\definecolor{baselinecolor}{gray}{.9}

\usepackage{pgf}
\usepackage{colortbl}
\usepackage{tipa}
\usepackage{tcolorbox}
\usepackage{rotating}
\usepackage[abs]{overpic}
\usepackage{makecell}
\usepackage{longtable}
\usepackage{tocloft}  %
\usepackage[english]{babel}
\usepackage{csquotes}
\usepackage{hyperref}
\usepackage[most]{tcolorbox}
\usepackage{fontawesome5}

\title{HY-WU (Part I): An Extensible Functional Neural Memory Framework and An Instantiation in Text-Guided Image Editing}

\author[]{Tencent HY Team}

\abstract{

Foundation models are transitioning from offline predictors to \emph{deployed systems} expected to operate over long time horizons. In real deployments, objectives are not fixed: domains drift, user preferences evolve, and new tasks appear after the model has shipped. This elevates \textbf{continual learning} and \textbf{instant personalization} from optional features to core architectural requirements. Yet most adaptation pipelines still follow a \emph{static weight paradigm}: after training (or after any adaptation step), inference executes a single parameter vector regardless of user intent, domain, or instance-specific constraints. This treats the trained or adapted model as a single point in parameter space. In heterogeneous and continually evolving regimes, distinct objectives can induce separated feasible regions over parameters, forcing any single shared update into compromise, interference, or overspecialization. As a result, continual learning and personalization are often implemented as repeated overwriting of shared weights, risking degradation of previously learned behaviors.
\vspace{-1em}

We propose \textbf{HY-WU (Weight Unleashing)}, a \textbf{memory-first} adaptation framework that shifts adaptation pressure away from overwriting a single shared parameter point. HY-WU implements \textbf{functional (operator-level) memory as a neural module}: a generator $g_{\phi}$ that synthesizes weight updates $\Delta\theta(x)$ on-the-fly from the instance condition, yielding instance-specific operators $f_{\theta+\Delta\theta(x)}$ \textbf{without test-time optimization}. Technically, HY-WU introduces: (i) an end-to-end on-the-fly training objective that optimizes the generator directly from downstream loss, avoiding checkpoint collection or reconstruction that hinder the scaling of prior hypernetwork-style approaches; (ii) a rank-anchored 2D parameter tokenization with factorized attention that scales weight generation to large backbones by exploiting architectural structure; and (iii) empirical and geometric analyses showing that gains arise from correct instance--parameter alignment rather than increased trainable capacity, and that generated updates organize into a semantically structured weight manifold.
\vspace{-1em}

We validate HY-WU as \textbf{Part I} of a series via a transformation-centric stress test: \textbf{text-guided image editing} (text-image-to-image), where objectives are often directional, mutually exclusive, and strongly instance-dependent. Under controlled conflict editing, static shared adaptation exhibits structural compromise, whereas HY-WU preserves directional specialization. Empirically, HY-WU shows strong competitive performance across both human and automatic evaluations: in pairwise human preference (GSB), HY-WU wins by large margins against all leading open-source editors (e.g., 67--78\% win rates against Step1X/Qwen/LongCat/FLUX) and also exceeds strong closed-source baselines (55.6\% vs.\ Seedream~4.5 and 55.5\% vs.\ GPT Image~1.5), while remaining close to the latest Nano-Banana series (47.6\% vs.\ Nano Banana~2 and 46.2\% vs.\ Nano Banana~Pro).
On public benchmarks, HY-WU ranks \#1 on GEdit-Bench and achieves \#2 on ImgEdit-Bench among open-source models.
In addition, on our internal WU-Eval, HY-WU also yields substantial gains across all dimensions.
\vspace{-1em}

These results motivate a broader perspective for foundational model and agentic AI design: adaptation becomes learning a mapping to a \emph{family} of parameter points, rather than optimizing a single shared solution (or a sequence of overwrites). With HY-WU, we advocate a complementary design path \textbf{beyond monolithic backbone scaling}: allocate capacity to \emph{structured, routable functional memory} that can specialize computation per instance without corrupting the shared base.

}

\date{\today}
\correspondence{Qinglin Lu and Kai Wang (\email{victorkwang@global.tencent.com})}

\metadata[Project page]{\url{https://tencent-hy-wu.github.io/}}

\begin{document}

\maketitle

\tableofcontents

\section{Introduction}
\label{section:intro}

Foundation models are increasingly deployed as persistent systems rather than static artifacts.
In real-world deployment, objectives evolve: user preferences drift, domains shift, and new tasks emerge after a model is released.
This operational reality makes \textbf{continual learning} and \textbf{instant personalization} central requirements.
A system should accumulate new competencies without degrading old ones, and adapt to a user, a session, or an instance with minimal delay and minimal data.
Crucially, these are not merely product features.
They are constraints on the \emph{memory interface} of the model, namely how new behaviors are stored and how they are applied at inference time.

\begin{figure*}[h]
    \centering
    \includegraphics[width=0.82\linewidth]{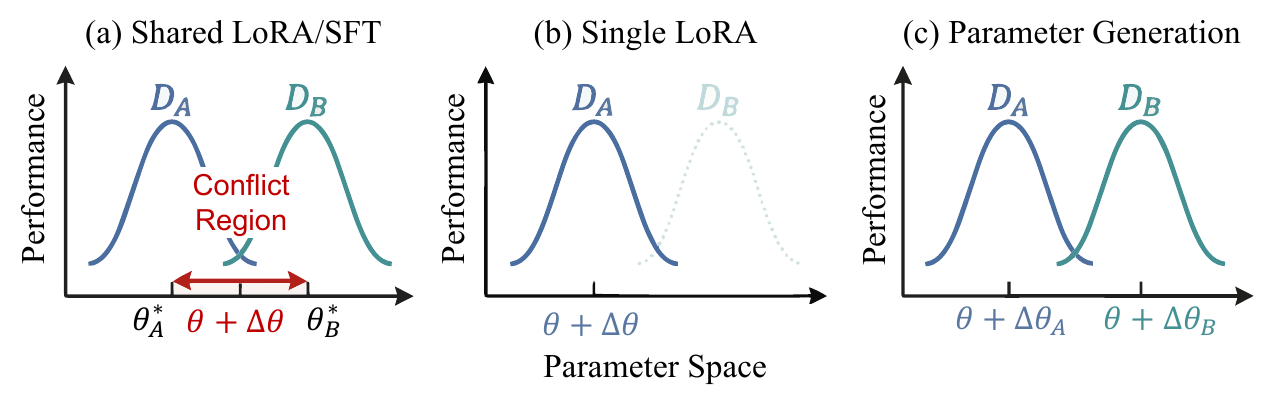}
    \caption{\textbf{Failure modes of static adaptation \textit{vs.} conditional parameter generation.}
    \textit{(a) Infeasible sharing.} When heterogeneous objectives induce separated feasible regions in parameter space, a single shared update (Shared LoRA or SFT) is forced into compromise, instability, or dominance by high-frequency modes.
    \textit{(b) Over-specialization.} Training a separate static adapter per domain avoids direct conflict but collapses into a narrow subspace and generalizes poorly under domain shifts.
    \textit{(c) Conditional generation.} A generator routes each instance to an update $\Delta\theta(x)=g_{\phi}(c(x))$, enabling inference over a family of parameter points rather than a single point. %
    }
    \label{fig:diagram}
\end{figure*}

Despite this shift, the dominant adaptation interface for foundation models remains \emph{static}.
Whether through supervised fine-tuning (SFT) or parameter-efficient methods such as adapters and LoRA, adaptation is typically realized as a single parameter update learned during training and then applied uniformly at inference.
Formally, given a pretrained backbone $f_{\theta}$, static adaptation learns one update $\Delta\theta_{\text{static}}$ and performs
\begin{equation}
\hat{y} = f(x; \theta + \Delta\theta_{\text{static}}), \quad \forall x .
\label{eq:static}
\end{equation}
This design assumes that heterogeneous deployment objectives can be compressed into a \emph{single point} in parameter space.
In narrow regimes, this compression can be adequate.
In heterogeneous and continually evolving regimes, we argue it becomes structurally brittle.

A useful way to see why is to treat adaptation as a feasibility problem in parameter space.
Each objective, domain, or user intent induces a region of parameter updates that achieve low loss for that objective.
Static adaptation seeks one point that works for all.
Figure~\ref{fig:diagram} illustrates two failure modes that follow from this single-point constraint, and both are common in practical deployments.
In Figure~\ref{fig:diagram}(a), the feasible regions for different objectives become separated.
Then a single shared update either does not exist or lies in a compromise region that partially satisfies none.
Empirically, such compromise can manifest as softened behavior, unstable training dynamics, or mode dominance under imbalance.
In Figure~\ref{fig:diagram}(b), one can abandon sharing and train separate static adapters per domain.
This avoids direct conflict, but it typically over-specializes: the adapter collapses into a narrow subspace tuned to one domain, and generalization suffers as soon as conditions shift.
Both failures share the same root cause: \emph{static adaptation commits inference to one fixed operator}.
From a continual-learning perspective, this means new behaviors must be written into the same point, making interference a structural consequence rather than an accidental training artifact.

\subsection{Memory Interfaces in Foundation Models: A Paradigm Shift}
\label{sec:memory_interfaces}

The above tension is not merely about optimization. It is about \emph{memory interfaces}.
A deployed system must store new information and skills, decide what to retain, and decide how to apply it at inference time.
Different memory interfaces store different kinds of information, expose different routing mechanisms, and induce different failure modes under heterogeneity.
In this report, we use ``memory'' in an adaptation sense: the mechanism by which a model accumulates and expresses new behaviors under evolving objectives.
We distinguish three interfaces that are most relevant to foundation-model adaptation.

\paragraph{(1) Static Parameter Memory.}
Fine-tuning and PEFT encode new behaviors by modifying shared parameters:
\begin{equation}
\hat{y} = f(x; \theta + \Delta\theta_{\text{static}}).
\end{equation}
This interface stores memory in the same parameter point that must serve all future inputs.
Continual learning therefore becomes repeated overwriting of shared weights.
When objectives conflict, interference is not a bug.
It is a consequence of single-point storage and single-operator inference.

\paragraph{(2) Context Memory (Activation + Retrieval).}
A common alternative is to store information externally and inject retrieved context into the forward pass:
\begin{equation}
\hat{y} = f(x, r(x); \theta),
\end{equation}
where $r(x)$ denotes retrieved items, summaries, or representations. Typically, activation memory resides inside the forward pass (context window/KV-cache), while retrieval memory pulls content from an external store.

This interface can be effective when the missing ingredient is \emph{information}, \eg, facts, exemplars, or user history that can be expressed as context.
However, the operator $f(\cdot;\theta)$ remains fixed.
When adaptation requires changing \emph{transformation rules} rather than adding context, retrieval alone does not directly modify the operator implementing those rules.
Importantly, this report does not benchmark retrieval-based memory baselines for image editing.
We treat retrieval memory as orthogonal and complementary, and return to it in the series roadmap.

\paragraph{(3) Functional Memory via Parameter Generation.}
A stronger interface treats memory as \emph{operator synthesis}.
Instead of retrieving content, the system generates an instance-conditioned parameter update:
\begin{equation}
\Delta\theta(x) = g_{\phi}(c(x)),
\qquad
\hat{y} = f(x; \theta + \Delta\theta(x)).
\label{eq:functional}
\end{equation}
Here $g_{\phi}$ is a neural module that \emph{implements memory}. It maps condition $c(x)$ to an operator update $\Delta\theta(x)$.
This is an explicit ``\textbf{memory as a neural network}'' formalization.
We call it \emph{functional memory}: memory is expressed as an operator-valued function, rather than as key--value items appended to activations.
Crucially, functional memory enables \emph{routing in weight space}: different conditions map to different regions of an update family, instead of forcing all behaviors into one shared update.

These interfaces imply different trade-offs for continual learning and instant personalization.
Static parameter memory overwrites a shared point.
Retrieval memory augments context but leaves the operator fixed.
Functional memory allows multiple operators to coexist as a conditional family, making it a natural substrate for fast instance-conditioned specialization and for continual adaptation under heterogeneous objectives.

If memory is implemented by a generator $g_{\phi}$ that outputs operator updates, then adaptation should be formulated as learning a \emph{conditional family} of parameter points together with an implicit routing rule.
Formally, learning $g_{\phi}$ induces a set
\begin{equation}
\mathcal{M}_{\phi} \;=\; \{ \Delta\theta(x) : x \sim \mathcal{D} \},
\end{equation}
which can be viewed as a structured manifold of updates over deployment conditions. Adaptation is then not ``find one update'' but ``learn a map from condition to update''. Consequently, when feasible regions are separated, the goal is no longer to force an intersection through a single shared update.
Instead, the system learns a conditional map that can route different instances to different regions in the update manifold.

That fundamentally reframes catastrophic interference (as in Figure~\ref{fig:diagram}) as a routing problem over a learned family rather than a repeated overwriting process.
In this view, the frozen backbone $\theta$ plays the role of stable invariants accumulated during pretraining, while $g_{\phi}$ functions as operator-valued working memory that synthesizes specialized transformations on demand.
A central empirical question then becomes whether the learned family is structured and semantically meaningful, rather than an unstructured collection of instance-wise perturbations.
Part~I addresses this question through conflict-controlled evaluations, alignment ablations, and weight-space geometry analyses.

\subsection{HY-WU (Part I):  A Stress-Test Instantiation in Text-Guided Image Editing}
\label{sec:intro_hywu}

We instantiate functional memory through \textbf{HY-WU (Weight Unleashing)}, a scalable framework for on-the-fly conditional generation of low-rank (LoRA) updates.
HY-WU synthesizes instance-conditioned adapter weights from hybrid image--instruction representations and injects them into a frozen backbone during the forward pass, producing instance-specific operators without test-time optimization.
A key practical barrier for prior hypernetwork-style approaches is reliance on pre-collected fine-tuned checkpoints and reconstruction losses.
HY-WU instead trains the generator end-to-end using only downstream task loss, enabling fully on-the-fly optimization and improving deployability.

We use \textbf{text-guided image editing} as a stress test because it makes heterogeneous, instance-dependent transformation rules unavoidable. This makes editing a strong testbed rather than a toy setting: conflicts are visually explicit, success criteria are assessable (faithfulness, structure preservation, controllability), and the need for instance-dependent operator shifts is unavoidable. Editing objectives are often directional and mutually exclusive, for example restoration vs. degradation, adding vs. removing content, or reversible domain transformations.
They are also strongly instance-dependent: the same instruction can require different transformation rules depending on image structure and semantics.
This is a procedural-memory regime: success depends on selecting and executing the right \emph{operator shift}, not merely retrieving additional context.
Accordingly, Part~I focuses on operator-level functional memory implemented via conditional LoRA generation under heterogeneous editing objectives, while full online CL protocols are an agenda item for later parts.

\paragraph{Contributions.}
This report makes the following contributions:
\begin{itemize}
    \item \textbf{Problem reframing for continual deployment.} We identify static adaptation as single-point inference in parameter space and characterize two structural failure modes under heterogeneity, infeasible sharing and over-specialization (Figure~\ref{fig:diagram}).
    \item \textbf{Memory-first formulation.} We formalize functional (operator-level) memory as ``memory as a neural network'' via conditional parameter generation, and clarify its relationship to static parameter memory and retrieval memory.
    \item \textbf{HY-WU system.} We propose a scalable conditional parameter generation framework that synthesizes LoRA updates on-the-fly from hybrid conditions, together with architectural and systems designs that make large-scale weight synthesis practical.
    \item \textbf{Mechanism and interpretability.} We provide controlled conflict studies and instance-alignment ablations showing that gains arise from correct condition--parameter alignment rather than capacity alone, and we analyze emergent semantic structure in generated parameter space, supporting the view of a structured conditional family of updates.
\end{itemize}

\begin{tcolorbox}[colback=gray!8, colframe=black!60]
\textbf{Positioning and scope.}
HY-WU should not be interpreted as a new image-editing backbone or as ``another dynamic LoRA'' method.
LoRA is used here as a structured \emph{operator interface} for functional memory, and text-guided image editing serves as a controlled, transformation-centric \emph{stress-test instantiation}. The broader claim is about a memory-first adaptation interface, namely routing over a conditional family of operator updates to reduce reliance on overwriting shared weights. 
\end{tcolorbox}

\section{Instantiation (Part I): Text-Guided Image Editing as a Procedural Functional-Memory Stress Test}
\label{sec:instantiation}

HY-WU is positioned as a memory-first adaptation framework whose core primitive is \emph{functional (operator-level) memory} implemented via conditional parameter generation (Section~\ref{sec:memory_interfaces}).
Part~I instantiates this framework in a concrete, transformation-centric regime: \textbf{text-guided image editing} (text-image-to-image, or TI2I).
We choose this setting not because HY-WU is inherently tied to diffusion models or image editing, but because editing provides a clean stress test where (i) objectives are often directional and mutually exclusive, and (ii) the correct behavior depends strongly on instance content.
Both properties sharply expose the limitations of single-point static adaptation (Figure~\ref{fig:diagram}) and make operator-level conditionality measurable via controlled conflicts, gradient analyses, and weight-space structure.

\subsection{Task Definition and Terminology}
\label{sec:editing_task}

In TI2I, each instance consists of an input image $I$ and an editing instruction $p$.
The goal is to produce an edited image $\hat{I}$ that (i) satisfies the instruction, (ii) preserves irrelevant content, and (iii) maintains structural coherence.
Throughout this report, we use a diffusion-based conditional generator as the backbone $f_{\theta}$, and we adapt it using LoRA as the structured interface for operator modification.

\paragraph{A memory-interface lens on ``personalization''.}
Section~\ref{sec:memory_interfaces} frames adaptation in terms of \emph{memory interfaces}, namely where new behavior is stored and how it is applied at inference time.
In generative vision, similar high-level ``personalization'' outcomes can be achieved through different interfaces: baking a concept into parameters, supplying concept evidence as conditioning context, or dynamically injecting transformation rules through operator updates.
We use this lens to clarify the relationship between \textbf{T2I personalization}, \textbf{TI2I} (image editing), and our notion of \textbf{personalized editing}.

From a \emph{functional} perspective, personalized editing can subsume many behaviors traditionally associated with T2I personalization (\eg, identity preservation, style transfer) as special cases.
Classical T2I personalization (\eg, DreamBooth) focuses on learning new \textit{concepts}, \textit{identities}, or \textit{styles} so that they can be generated from text prompts.
This can be viewed as expanding the model’s representational space: enabling the backbone to synthesize a previously unseen concept.
The adaptation unit is persistent, relying on static parameter memory to recall the subject.
In contrast, we define personalized editing as the instance-conditioned specialization of transformation behavior in text-guided image editing.
When the editing rule enforces identity consistency, style coherence, or subject-specific constraints, the operator must simultaneously modify certain attributes while maintaining core characteristics invariant.
This distinguishes personalized editing from generic image editing: generic editing typically executes a fixed operator (the base model), whereas personalized editing requires the \emph{transformation rule} itself to specialize under instance-specific invariance constraints.

Under this view, certain behaviors linked to T2I personalization (\eg, identity consistency) can be realized as operator specialization conditioned on instance-specific signals:
if the target identity is already present in the input image and the instruction demands consistency, the required behavior reduces to operator specialization under invariance constraints.
Thus, many personalization effects typically achieved through concept injection in T2I personalization can also emerge through instance-conditioned operator shifts in image editing.

The key distinction lies in the adaptation unit:
T2I personalization expands representational memory for a subject, whereas personalized editing specializes the transformation rule applied to a given instance.

\vspace{0.5em}
\begin{tcolorbox}[colback=gray!6, colframe=black!55, title=\textbf{Terminology guardrail}]
\begin{itemize}
    \item \textbf{T2I personalization} expands a model’s representational memory and concept vocabulary by learning new concepts, identities, or styles, enabling subject generation from text prompts.
    (In the memory taxonomy of Section~\ref{sec:memory_interfaces}, this is typically realized as \emph{static parameter memory}.)
    
    \item \textbf{TI2I image editing} conditions on a source image, where identity/style signals are often already present in the instance input and can be provided through conditioning context (and optionally retrieved references), while the operator may remain fixed.
    (This aligns with \emph{context/activation-level interfaces}; it does not, by itself, imply persistent memory across time.)

    \item \textbf{Personalized editing} refers to instance-conditioned operator specialization for transformation rules in text-guided image editing, adapting how existing image(s) are modified under procedural constraints.
    (This corresponds to \emph{functional memory} implemented as operator synthesis.)
\end{itemize}
\end{tcolorbox}
\vspace{0.5em}

\subsection{Why Editing is a Procedural-Memory Stress Test}
\label{sec:editing_stresstest}

Text-guided editing is a particularly sharp stress test for static adaptation because editing objectives are often \emph{directional} and can be \emph{mutually exclusive}.
For example, transformations like ``restore'' \textit{vs.} ``age'' or ``blur'' \textit{vs.} ``deblur'' often correspond to incompatible operations.
When such conflicting objectives are mixed during adaptation, a single shared update is forced to compromise, yielding softened edits, unstable behavior, or dominance by high-frequency modes.
This aligns with Figure~\ref{fig:diagram}(a): different objectives can induce separated feasible regions in parameter space, making single-point sharing structurally ill-posed.

A second stressor is \emph{instance dependence}.
Even under identical instruction text, the optimal transformation varies drastically based on the input image's content, layout, and semantics. 
Editing is therefore not well modeled by a globally fixed adapter that applies the same operator shift to all instances.
This aligns with Figure~\ref{fig:diagram}(b): static domain adapters can over-specialize to the training distribution and suffer from poor generalization under instance shifts.
In other words, editing naturally emphasizes \emph{procedural memory}: a system must reliably execute the correct transformation rule conditioned on the specific instance.

These underlying properties motivate our use of TI2I editing as a stress-test instantiation for functional memory. 
In this regime, operator-level conditionality is not merely an aesthetic preference.
It is a necessary representational tool to prevent mutually exclusive objectives from being compressed into a single operator.

\subsection{Mapping the Memory-First Framework to the Editing Instantiation}
\label{sec:mapping_table}

Part~I is designed so that each abstract component in Section~\ref{section:intro} has a concrete counterpart in the TI2I system, and each claim is related to a measurable diagnostic in our experiments.
Table~\ref{tab:concept_mapping} summarizes this mapping.

\begin{table*}[t]
\centering
\small
\begin{tabular}{p{0.22\linewidth} p{0.35\linewidth} p{0.37\linewidth}}
\toprule
\textbf{Framework concept} & \textbf{Instantiation in Part I (TI2I)} & \textbf{Where it is tested / evidenced} \\
\midrule
Stable invariants \newline (long-term knowledge) &
Frozen diffusion backbone $f_{\theta}$ (pretrained competence preserved during adaptation) &
Static baselines show compromise or over-specialization under conflict (qualitative and quantitative), motivating preservation of the shared base \\
\addlinespace
Static parameter memory \newline (single-point adaptation) &
SFT / Shared LoRA producing a fixed update $\Delta\theta_{\text{static}}$ &
Conflict-controlled editing reveals compromise under sharing; gradient conflict analyses characterize incompatibility \\
\addlinespace
Functional memory \newline (operator-level) &
Generator $g_{\phi}$ synthesizing LoRA updates $\Delta\theta(x)$ per instance &
Instance-alignment ablations isolate routing as mechanism (Table~\ref{tab:instance_ada}); weight-space structure analyses probe semantic organization \\
\addlinespace
Routing signal / condition &
Hybrid condition $c(x)$ from image and instruction features &
Instance dependence: same instruction requires different operator shifts depending on content; ablations show that misalignment collapses gains \\
\addlinespace
Conditional family of operators &
$\{\theta + \Delta\theta(x)\}$ induced by $g_{\phi}$ (update family over instances) &
Emergent parameter-space neighborhoods align with semantic similarity; clustering and local neighborhood consistency in generated parameters \\
\addlinespace
Failure mode under heterogeneity &
Separated feasible regions and incompatible gradients &
Controlled conflicts (e.g., restoration vs.\ aging) and gradient-level conflict evidence \\
\bottomrule
\end{tabular}

\caption{\textbf{Concept-to-implementation mapping for Part~I (TI2I).}
We map the memory-first framing to concrete modules, and we indicate where each concept is validated in Part~I.}
\label{tab:concept_mapping}

\end{table*}

Two scope clarifications are essential. First, Part~I focuses on \emph{operator-level} functional memory and does not benchmark retrieval-augmented editing or agent-style external memory baselines.
Second, LoRA is used as a \emph{structured operator interface}, not as a claim that the framework is inherently LoRA-specific.

Later parts of the HY-WU series will broaden both the application domain (\eg, longer-horizon multimodal settings) and the memory substrate (\eg, integration with retrieval memory).

\subsection{Baselines and Diagnostic Controls Under the Memory Lens}
\label{sec:baselines_memorylens}

Our evaluation is designed to separate three confounders that frequently blur conclusions in adaptation studies: (i) trainable parameter count, (ii) modular isolation, and (iii) conditional routing.
Accordingly, we compare:
(i) \textbf{SFT} and \textbf{Shared LoRA} as single-point static parameter memory baselines,
(ii) \textbf{Single LoRA} (one adapter per objective/domain) as an isolation baseline that avoids direct conflict but risks over-specialization,
and (iii) \textbf{HY-WU} as functional memory via instance-conditioned operator synthesis.
To probe mechanism rather than merely behavioral performance, we further design diagnostic controls that preserve parameter count while removing instance–update correspondence (detailed in Section~\ref{sec:instance_adaptivity}). 
These controls allow us to causally attribute performance gains to conditional routing rather than raw capacity or modular isolation.

\subsection{Task-Specific Background: Editing, Controllability, and Personalization in Diffusion Models}
\label{sec:rw_editing_local}

Text-guided image editing has advanced rapidly with diffusion models, with a broad spectrum of approaches targeting controllability, faithfulness, and structure preservation.
A common theme is to treat editing as conditional generation guided by text, sometimes aided by inversion or intermediate constraints, \eg, SDEdit~\citep{meng2021sdedit}, prompt-to-prompt editing~\citep{hertz2022prompt}, null-text inversion~\citep{mokady2023nulltext}, and instruction-following editing methods~\citep{brooks2023instructpix2pix,Zhang2023MagicBrush}.

More advanced models~\citep{wang2025seededit,deng2025emerging,labs2025flux1kontextflowmatching,wu2025omnigen2,hidreami1technicalreport,wu2025qwen,openai2025gptimage15,liu2025step1x-edit,cao2025hunyuanimage,flux2,LongCatImage,team2026firered}, with scaled capabilities, have emerged as a fundamental component of modern artificial intelligence, achieving remarkable text-guided image editing performance. Many introduced explicit structural controls, such as conditioning on edges, depth, segmentation, or reference features (\eg, ControlNet~\citep{zhang2023adding}) or adapters~\citep{ye2023ipadapter,tao2025instantcharacter}.
These methods improve controllability and local consistency, but typically rely on a fixed backbone operator, with adaptation (if any) implemented as a static fine-tuning step.

Personalization in diffusion models often targets new concept acquisition and identity preservation, traditionally via DreamBooth-style fine-tuning or parameter-efficient variants (\eg, LoRA-based personalization, textual inversion).
These settings highlight a practical tension between data efficiency, overfitting, and interference when multiple concepts or styles must be supported in one system~\citep{gal2022IMAGE,ruiz2023dreambooth,qian2025omni}.
Our Part~I setting differs in emphasis: we treat editing personalization primarily as \emph{procedural specialization} for transformation rules, where conflicts between objectives are explicit and instance-level conditionality is central.
This makes TI2I editing a clean arena to test whether a functional-memory interface can preserve directional behaviors without collapsing heterogeneous objectives into a single static compromise.

\section{HY-WU Implementation for Text-Guided Image Editing}
\label{sec:implementation}

\begin{figure*}[h]
    \centering
    \begin{subfigure}[t]{0.32\textwidth}
        \centering
        \includegraphics[width=\textwidth]{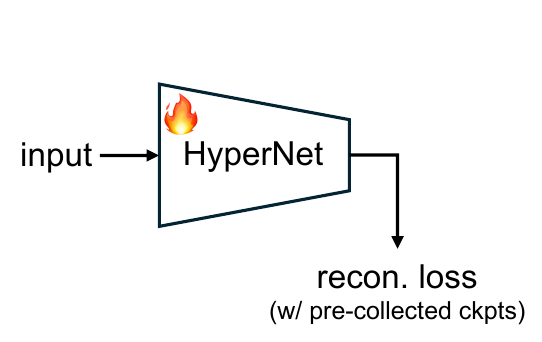}
        \caption{Learning to reconstruct parameters of pre-collected checkpoints.}
        \label{fig:only_recon}
    \end{subfigure}
    \hfill
    \begin{subfigure}[t]{0.32\textwidth}
        \centering
        \includegraphics[width=\textwidth]{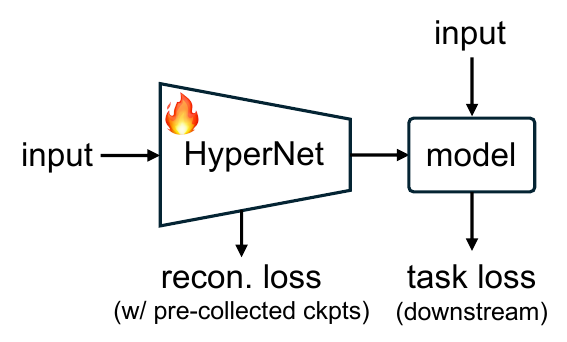}
        \caption{Learning to reconstruct parameters and downstream task loss as auxiliary.}
        \label{fig:recon_and_downstream}
    \end{subfigure}
    \hfill
    \begin{subfigure}[t]{0.32\textwidth}
        \centering
        \includegraphics[width=\textwidth]{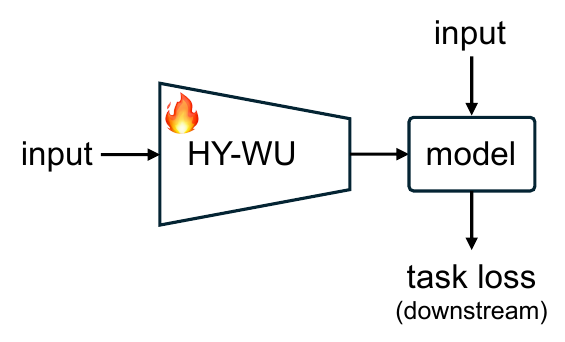}
        \caption{On-the-fly optimization with only downstream task loss.}
        \label{fig:simple_wu}
    \end{subfigure}
\caption{Comparison of training paradigms for hypernetwork-based parameter generation.
\textbf{(a)} Learning to reconstruct parameters from pre-collected checkpoints via reconstruction loss.
\textbf{(b)} Learning from pre-collected checkpoints with additional downstream task loss as auxiliary supervision.
\textbf{(c)} Our approach: on-the-fly optimization of the parameter generator using only downstream task loss, without relying on pre-collected checkpoints.
}

\label{fig:hypernet_vs_follow_vs_wu}
\end{figure*}

This section specifies how HY-WU operationalizes functional memory for TI2I editing.
Given an instance $x=(I,p)$, HY-WU computes a condition embedding $c(x)$, generates LoRA updates $\Delta\theta(x)$, inserts them into the frozen backbone, and executes the instance-specific operator $f_{\theta+\Delta\theta(x)}$.
We focus on designs that make conditional weight synthesis scalable for modern large backbones, including the tokenization of heterogeneous parameters, architecture-aware attention, stable initialization, and distributed training strategies for long parameter sequences.

\subsection{Prerequisite: From Checkpoint-Supervised Generation to On-the-Fly Training}
\label{sec:onthefly_motivation}

Parameter generation has a long history in hypernetwork-style approaches~\citep{ha2017hypernetworks,stanley2009hypercube}.
Recent efforts improve scalability by learning from pre-collected checkpoints via reconstruction losses, sometimes augmented with downstream objectives~\citep{schurholt2021self,wang2024neural,jin2024conditional,schurholt2024sane,li2024tina,ruiz2024hyperdreambooth,peebles2022learning,schurholt2022hyper,wang2025recurrent,liang2025draganddrop,liu2026shinescalableincontexthypernetwork}.
However, checkpoint collection introduces practical constraints: storing large numbers of adapters, curating a target distribution of ``desired weights'' and paying I/O costs during training.

HY-WU adopts a different regime.
We treat the generator as \emph{functional memory} trained \emph{on the fly} directly through the downstream editing loss, avoiding explicit weight supervision, as illustrated in Figure \ref{fig:hypernet_vs_follow_vs_wu}.
This design choice is essential for scaling conditional operator synthesis when the number of conditions, objectives, or user contexts grows, where maintaining a checkpoint bank becomes increasingly expensive and brittle.
In HY-WU, the backbone remains frozen and gradients propagate through the generated weights into the generator, enabling the system to learn the mapping from conditions to operators directly.

\subsection{System Overview}
\label{sec:hywu_overview}

\begin{figure*}[t]
    \centering
    \centering
    \includegraphics[width=0.8\textwidth]{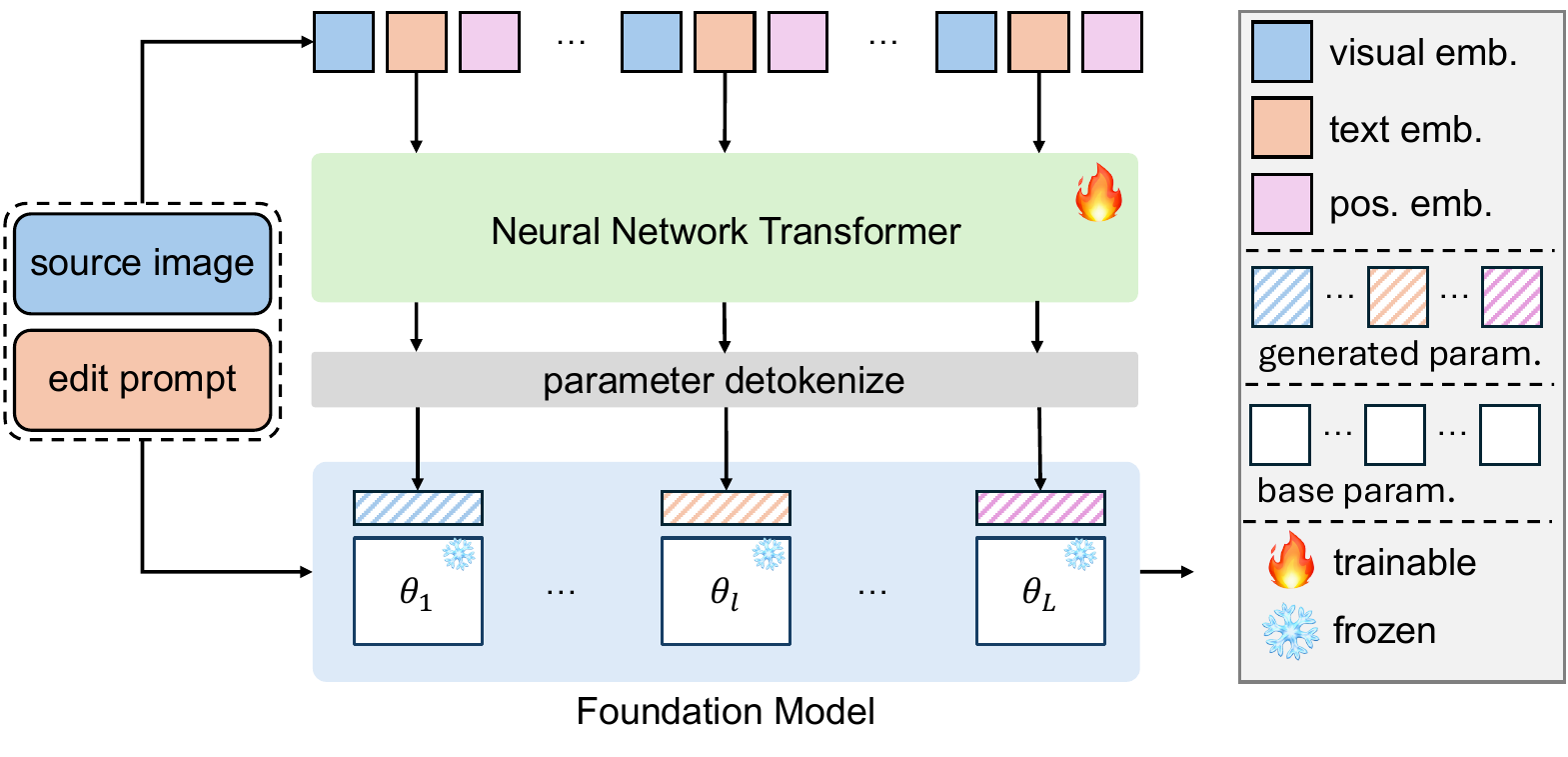}
    \caption{Overview of the HY-WU pipeline. The framework extracts conditions from the source image and edit prompt, which are processed by a trainable Neural Network Transformer to synthesize instance-specific parameter tokens. These tokens are then detokenized into LoRA adapters and integrated into a frozen foundation model with $\theta_1\cdots\theta_L$, where $\theta_l$ indicates $l$-th layer. The entire pipeline is optimized end-to-end, where the generator is updated via backpropagation of diffusion loss.}
    \label{fig:pipeline}
\end{figure*}

As shown in Figure~\ref{fig:pipeline}, HY-WU generates instance-specific LoRA adapters through three stages.
First, it extracts hybrid condition features from the input image and instruction prompt.
Second, a transformer-based parameter generator synthesizes these conditions into a structured sequence of parameter tokens.
Third, the tokens are detokenized into LoRA matrices and injected into the foundation model.
During training, the foundation model remains frozen and is executed with the generated adapters.
During backpropagation, gradients flow through the LoRA-injected forward pass to update only the generator (and lightweight conditioning modules, if any).

\subsection{Hybrid Condition Extraction}
\label{sec:condition_extraction}

A key requirement for instance-conditioned operator synthesis is that the condition $c(x)$ must encode both \emph{what} edit is requested and \emph{where / how} it should apply given image content.
We therefore construct a hybrid condition representation by combining image and text features.
In our default implementation, we adopt a vision-language encoder (SigLIP2~\citep{tschannen2025siglip} in our experiments) and compute:
\begin{equation}
U = \Big[\mathrm{Enc}_{\mathrm{img}}(I;\vartheta)\ \oplus\ \mathrm{Enc}_{\mathrm{text}}(p;\vartheta)\Big],
\end{equation}
where $\oplus$ denotes concatenation and $\vartheta$ denotes the encoder parameters.
The resulting $U$ serves as the conditioning input to the parameter generator via cross-attention (Section~\ref{sec:nnt}).
This design is modular: the encoders can be replaced by alternative multimodal representations depending on the application, without changing the parameter-token interface.

\subsection{Rank-Anchored 2D Parameter Tokenization}
\label{sec:tokenization}

\begin{wrapfigure}{r}{0.6\textwidth}
    \centering
    \vspace{-15pt} %
    \includegraphics[width=0.58\textwidth]{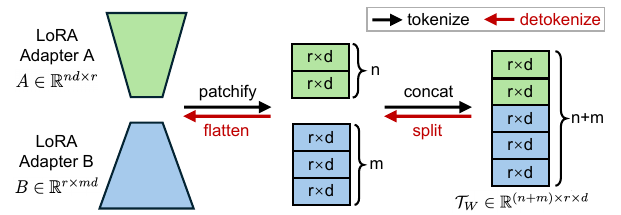}
    \vspace{-8pt}
    \caption{Parameter tokenization and detokenization. LoRA adapters are reorganized into a unified tensor $\mathcal{T}_{w}$. The red path reverses this process to reconstruct parameters from a tensor.}
    \vspace{-4pt}
    \label{fig:tokenize}
\end{wrapfigure}

Generating LoRA parameters for large backbones poses a structural challenge:
adapter matrices of backbone models across layers have heterogeneous input and output dimensions, while Transformer-based parameter generators require tokens of uniform shape. 
A naive flattening of each matrix would destroy its two-dimensional structure and entangle rank and channel semantics.
Although layer dimensions vary, the LoRA rank $r$ is fixed across modules.
This provides a natural anchor dimension shared by all adapters.
We therefore treat the rank dimension as a stable axis and reorganize the remaining spatial dimensions into fixed-length segments.

As shown in Figure~\ref{fig:tokenize}, for a backbone weight matrix 
$W \in \mathbb{R}^{d_{\mathrm{in}} \times d_{\mathrm{out}}}$, 
we compute $d = \gcd(d_{\mathrm{in}}, d_{\mathrm{out}})$ and decompose
\[d_{\mathrm{in}} = n d, \qquad d_{\mathrm{out}} = m d.\]
Given LoRA adapters $A \in \mathbb{R}^{nd \times r}$ and $B \in \mathbb{R}^{r \times md}$,
we partition them along the $d$ dimension and reorganize them into 
\[\mathcal{T}_{W} \in \mathbb{R}^{(n+m) \times r \times d}.\]
Each $r \times d$ slice forms a \emph{parameter token} with a consistent shape across layers.
Within each backbone layer, tokens from all adapted modules 
(\eg, attention projections) are concatenated into
\[\mathcal{T}_{L} \in \mathbb{R}^{s \times r \times d},\]
where $s$ is the number of tokens in that layer.
Across $l$ layers, the entire parameter set is represented as
\[\mathcal{T} \in \mathbb{R}^{l \times s \times r \times d}.\]

The inverse process detokenizes the generated tensor back into LoRA matrices $\{A,B\}$ and injects them into the backbone.
Intuitively, each parameter token ($r\times d$) encapsulates a small local slice of channels (of size $d$) in the original weight matrix, and $r$ dimension captures the full set of low-rank update directions.
In other words, a token preserves how a local channel segment interacts with each rank component of the LoRA matrix.

This rank-anchored tokenization has two benefits: i). It standardizes token shape while preserving the intrinsic 2D structure of adapters, despite heterogeneous layer dimensions across layers.
ii). Explicit indexing along the layer ($l$), token ($s$), rank ($r$), and segment ($d$) axes preserves architectural locality, allowing the subsequent generator (Section~\ref{sec:nnt}) to employ factorized attention patterns that respect layer and module boundaries.

\subsection{Neural Network Transformer (Parameter Generator)}
\label{sec:nnt}

\begin{figure*}[h]
    \centering
    \includegraphics[width=0.8\textwidth]{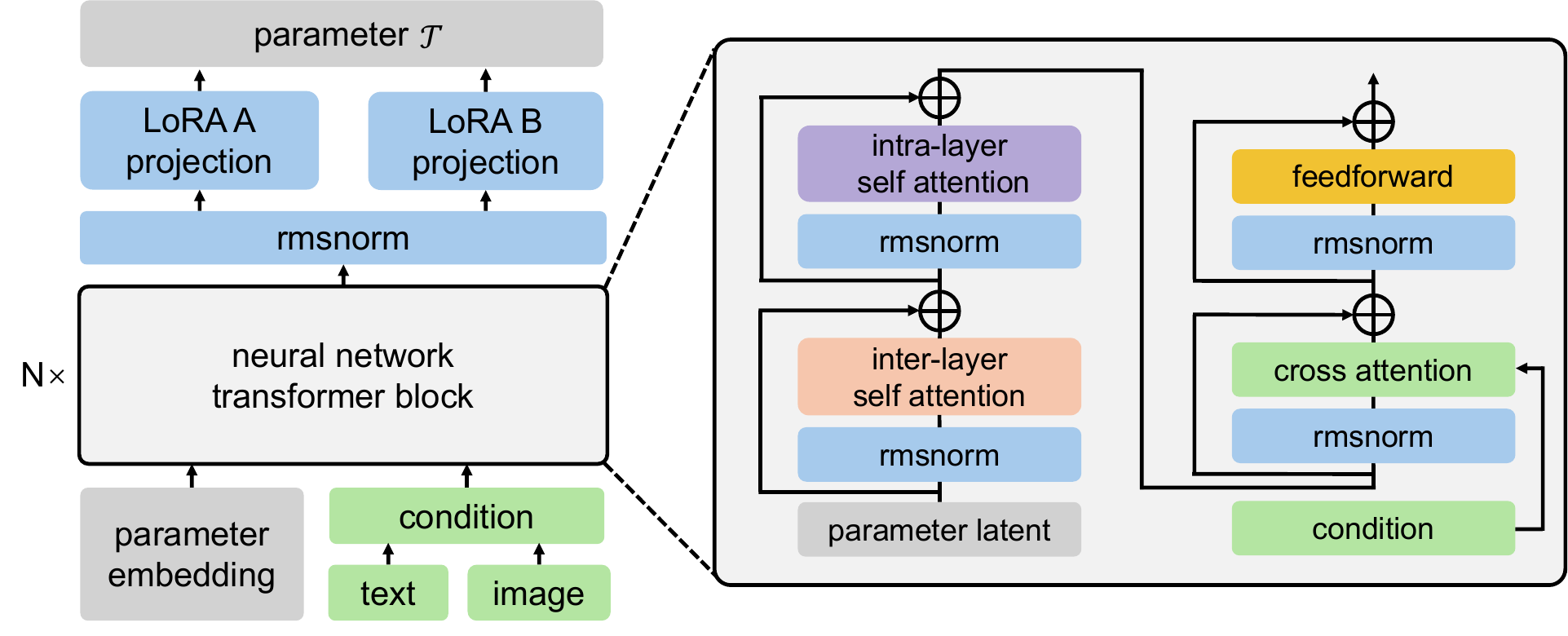}
    \caption{Architecture of the Neural Network Transformer. The left panel illustrates the overall pipeline, where parameter embeddings and extracted conditions (text/image) are processed through $N$ transformer blocks to generate LoRA parameters $\mathcal{T}$. The right panel details the internal structure of each block, featuring factorized self-attention to capture structural correlations, and cross-attention for condition injection. The final LoRA $B$ projection is zero-initialized to ensure training stability.}
    \label{fig:model}

\label{fig:pipeline_and_arch}
\end{figure*}

HY-WU uses a transformer-based parameter generator, which we refer to as the \textbf{Neural Network Transformer} (NNT), to map conditions $U$ to parameter tokens $\mathcal{T}$.
NNT is designed to model correlations in parameter space while remaining scalable for long sequences.

\paragraph{Factorized self-attention.}
Naive self-attention over the flattened sequence of length $l \cdot s \cdot r$ is quadratic and becomes prohibitive.
We therefore adopt factorized attention~\citep{Child2019GeneratingLS,9710415}, decomposing attention into two structured operations over $\mathcal{T} \in \mathbb{R}^{l \times s \times r \times d}$:

\begin{itemize}
    \item \textbf{Intra-layer attention:} model dependencies among all tokens within a layer by reshaping to $\mathbb{R}^{l \times (s\cdot r) \times d}$ and attending along the $(s\cdot r)$ dimension. This captures correlations among adapted modules within the same backbone block.
    \item \textbf{Inter-layer attention:} model dependencies across depth by reshaping to $\mathbb{R}^{(s\cdot r) \times l \times d}$ and attending along the $l$ dimension. This captures how functionally corresponding modules evolve across layers.
\end{itemize}

This design reflects two empirical priors: parameters within a layer are strongly coupled by the block structure, while cross-layer relationships are more hierarchical and role-consistent (\eg, $W_q$ across depths).
To preserve architectural indices, we apply independent Rotary Positional Embeddings (RoPE)~\citep{su2024roformer} to the layer index ($l$), token index ($s$), and rank index ($r$), enabling the generator to remain aware of each token's structural role.

\paragraph{Injecting conditions into parameter priors.} We initialize the latent with learnable parameter embeddings that serve as a structural prior. It is then modulated by conditions $U$ via cross-attention, transforming the task-agnostic initialization into instance-specific parameter tokens.

\paragraph{Stability via zero initialization.}
We incorporate a zero-initialization scheme to ensure a stable transition from the pretrained backbone at the start of training.
Concretely, we use separate output projections for the tokens corresponding to LoRA matrices $\{A,B\}$, and initialize the projection for $B$ to zero.
This ensures the initial generated adapter produces a negligible effect, so training begins from the original pretrained operator and gradually learns instance-conditioned deviations.

\subsection{On-the-Fly End-to-End Training}
\label{sec:training}

HY-WU is trained end-to-end directly through the downstream editing loss, without reconstructing pre-collected adapter checkpoints.
Let $\mathcal{L}_{\mathrm{edit}}(\hat{I}, I, p)$ denote the editing objective (e.g., diffusion denoising loss under conditional guidance).
Training proceeds as:
\begin{equation}
U \leftarrow \mathrm{Enc}(I,p), \qquad
\Delta\theta(x) \leftarrow g_{\phi}(U), \qquad
\hat{I} \leftarrow f_{\theta+\Delta\theta(x)}(I,p),
\qquad
\phi \leftarrow \phi - \eta \nabla_{\phi}\mathcal{L}_{\mathrm{edit}}.
\end{equation}
The backbone parameters $\theta$ remain frozen, and gradients propagate through the generated LoRA weights back to $\phi$.
This on-the-fly optimization has two practical consequences.

\paragraph{Training efficiency.}
By eliminating reconstruction losses and weight-target datasets, we remove the prerequisite of training, storing, and loading large-scale LoRA checkpoints.
This reduces storage overhead and I/O bottlenecks, which become dominant at scale when the number of conditions grows.

\paragraph{Unconstrained optimization in update space.}
Checkpoint-supervised generation constrains the generator to mimic a target distribution of updates, which can bias learning toward narrow regions of parameter space.
In contrast, direct optimization under $\mathcal{L}_{\mathrm{edit}}$ allows the generator to discover updates that are optimal for the downstream objective, without being restricted to the geometry of a pre-collected adapter bank.
We empirically investigate this effect via parameter-space analyses (\eg, Figure~\ref{fig:train_vs_pg_tsne}) and alignment ablations (Table~\ref{tab:instance_ada}) that isolate the role of instance-conditioned routing.

\subsection{Infrastructure and Systems Optimizations}
\label{sec:infrastructure}

Parameter generation introduces a systems challenge that is often under-emphasized: the generated parameter sequence can be extremely long, and naive training incurs substantial memory overhead.
We implement HY-WU with distributed strategies designed for long sequences and large backbones.

\paragraph{Distributed training.}
We adopt a hybrid parallelization strategy.
We apply Data Parallelism (DP) to the frozen image and text encoders, which have modest memory overhead.
For the diffusion backbone and the parameter generator, we use FSDP2~\citep{Zhao2023PyTorchFE} to shard parameters, gradients, and optimizer states across GPUs.
We additionally apply gradient checkpointing across layers to reduce activation memory.
To handle long sequences, we integrate sequence parallelism: DeepSpeed-Ulysses~\citep{Jacobs2023DeepSpeedUS} is used for the foundation model, while DSP~\citep{Zhao2024DSPDS} is used for the parameter generator.

\paragraph{Low-level Acceleration.}
For foundation models utilizing Mixture-of-Experts (MoE), kernel efficiency is a critical determinant of throughput. We optimize these components using FlashInfer~\citep{ye2025flashinfer} for inference-time acceleration and Triton-based~\citep{tillet2019triton} kernels for training efficiency. To further enhance performance, both the parameter generator and the foundation models are compiled with torch.compile~\citep{ansel2024pytorch}. Together, these optimizations significantly reduce latency and maximize throughput across both training and inference stages.

\section{Model Performance}
\label{sec:main_performance}

\begin{tcolorbox}[colback=gray!8, colframe=black!60, title=\textbf{Takeaways}]
\begin{itemize}
    \item \textbf{Human preference (GSB):} In pairwise comparisons, HY-WU decisively outperforms leading open-source editors (e.g., 67--78\% win rates vs.\ Step1X/Qwen/LongCat/FLUX) and also exceeds strong closed-source baselines (55.6\% vs.\ Seedream~4.5 and 55.5\% vs.\ GPT Image~1.5), while remaining competitive against the latest Nano-Banana series (Section~\ref{sec:humaneval}).
    \item \textbf{Automatic evaluation (WU-Eval and public benchmarks):} On our internal WU-Eval suite with a VLM judge, integrating HY-WU consistently improves alignment, consistency, structure, and quality over the corresponding backbone. On public benchmarks, HY-WU ranks \#1 among all open-source models on GEdit-Bench (EN/CN overall and semantic consistency) and \#2 among public models on ImgEdit-Bench (Section~\ref{sec:autobench}).
    \item \textbf{Universality and scaling trends:} HY-WU improves performance across two backbone families (a native unified multimodal model and an MMDiT diffusion transformer). We also demonstrate that increasing generator capacity and LoRA rank yields clearly improving trends (i.e., ``positive scaling law of functional memory") (Section~\ref{sec:ablation}).
\end{itemize}
\end{tcolorbox}

\subsection{Experimental Setups}
\label{sec:exp_setup}

\begin{table*}[h]

\centering %
\tablestyle{11pt}{1.2}

\begin{minipage}{0.4\textwidth} %
\centering
\begin{tabular}{c|c}
\textbf{NNT configuration} & \textbf{value} \\
\noalign{\vspace{1pt}}
\shline
\noalign{\vspace{1pt}}
parameter count & 8.11B \\
number of layers & 24 \\
number of heads & 32 \\
hidden dimension & 4096 \\
intermediate dimension & 16384
\end{tabular}
\end{minipage}
\hspace{0.01\textwidth} %
\begin{minipage}{0.4\textwidth}
\centering
\begin{tabular}{c|c}
\textbf{generated LoRA configuration} & \textbf{value} \\
\noalign{\vspace{1pt}}
\shline
\noalign{\vspace{1pt}}
parameter count & 0.72B \\
foundation model layer $l$ & 32\\
token per layer $s$ & 24 \\
LoRA rank $r$ & 16 \\
dimension $d$ & 128 \\
\end{tabular}
\end{minipage}

\caption{\textbf{Configuration for Neural Network Transformer (NNT) and generated LoRA on HY-Image-3.0-Instruct (80B-A13B).}}
\label{tab:configuration}
\end{table*}

We integrate the HY-WU framework with HY-Image-3.0-Instruct, an 80B (13B active) parameter native multimodal foundation model. For weight generation, we employ an 8.11B parameter transformer to produce 0.72B rank-16 LoRA parameters for all linear modules. Model details can be found in Table \ref{tab:configuration}.

\paragraph{Evaluation Protocol.}
In this report, we constructed a comprehensive multimodal benchmark, covering both single- and multi-image editing. The two tracks comprise $346$ and $64$ bilingual (Chinese-English) edit pairs, covering 60 image editing sub-tasks and diverse scenarios and objects.
Based on this benchmark, we conduct a comprehensive evaluation of state-of-the-art image editing models.
The participants include closed-source systems such as Seedream 4.5~\citep{seedream2025seedream}, GPT Image 1.5~\citep{openai2025gptimage15}, Gemini-3.0 Pro Image~\citep{google2025nanobanana} (abbreviated as Nano Banana Pro),
and open-source models such as HY-Image-3.0-Instruct~\citep{cao2025hunyuanimage}, Qwen-Image-Edit~\citep{wu2025qwen}, FLUX.2~\citep{flux2}, Step1X-Edit~\citep{liu2025step1x-edit}, and LongCat-Image-Edit~\citep{LongCatImage}.

\subsection{Human Evaluation} %
\label{sec:humaneval}

\paragraph{GSB results.} 
To comprehensively assess the perceptual quality, we conduct a human evaluation (GSB) against strong open- and closed-source baselines.
In each pairwise comparison, two model outputs for the same input are presented simultaneously in randomized order, and annotators are required to select their preferred result.
This evaluation protocol eliminates presentation bias and aligns closely with human perception.
For each model pair, annotators label the comparison as:
\textbf{G} (good), 
\textbf{S} (same), or 
\textbf{B} (bad).
Win rate is computed as $(G + 0.5S)/(G + S + B)$.
This formulation accounts for ties while preserving symmetric comparison.

\begin{figure*}[h]
    \centering
    \includegraphics[width=0.55\textwidth]{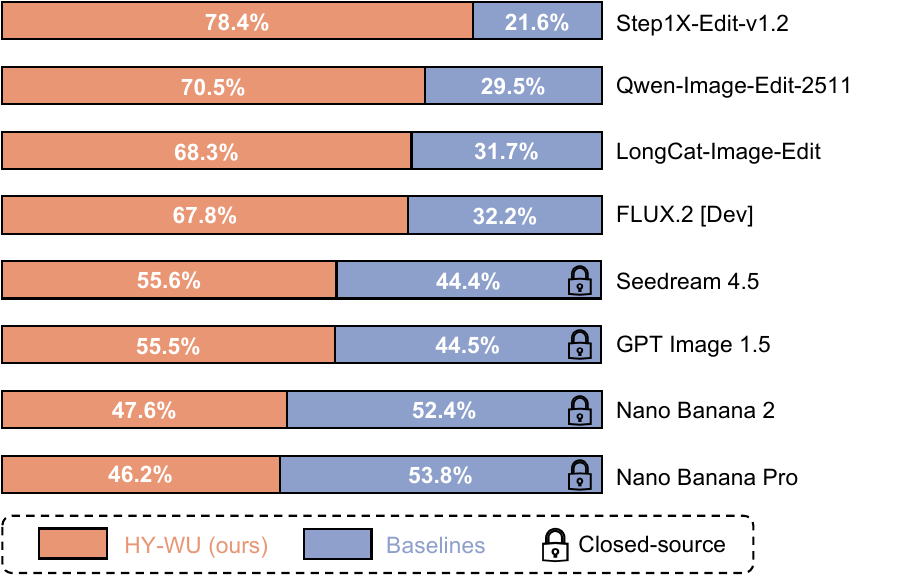}
    \caption{\textbf{The GSB win-rate comparison of our HY-WU against other baseline models.} Each bar reports the pairwise win rate of HY-WU against the corresponding baseline (rates sum to 100\%). 
    HY-WU demonstrates a significant performance advantage, particularly against established open-source competitors.
    Notably, HY-WU also achieves a clear lead over two closed-source competitors (marked with a lock icon), \ie, Seedream-4.5-2K (55.6\%) and GPT-Image-1.5 (55.5\%), while remaining highly competitive against the state-of-the-art Nano-Banana series, proving its effectiveness in complex image editing tasks.
    }
    \label{fig:humaneval}
\end{figure*}

As illustrated in Figure~\ref{fig:humaneval}, HY-WU substantially outperforms leading open-source models, and remain competitive with top-tier closed-source commercial systems.
While Nano Banana 2 and Nano Banana Pro achieve slightly higher overall scores (52.4\% and 53.8\%, respectively), the margin remains modest.
Given that these commercial systems are likely trained with substantially larger-scale backbones and proprietary data, the modest performance gap suggests that our operator-level conditional adaptation remains effective even under more constrained model scale.

\subsection{Quantitative Results}
\label{sec:autobench}
In addition to human evaluation, we also develop an automatic benchmarking system, namely \textit{WU-Eval}, based on VLM Judge (Qwen-3-VL-32B~\citep{qwen3}), which is essential for large-scale evaluation and provides fast and stable feedback.
The scoring process is broken down into fine-grained visual-question-answer evaluating $4$ dimensions, \ie, instruction alignment, consistency, structure, and quality, ensuring the stability and interpretability of the benchmarking system.
Following common practice, we also include GEdit-Bench~\citep{liu2025step1x-edit} and ImgEdit Benchmark~\citep{ye2025imgedit} for comprehensive evaluation.

\paragraph{GEdit-Bench.}
We evaluate our model on GEdit-Bench~\citep{liu2025step1x-edit}, which provides analyzes of semantic consistency and perceptual quality.
As the evaluation scores from GPT-4o reported by original paper exhibit fluctuation over time, we only evaluate the results using Qwen-2.5-VL-72B~\citep{qwen25}.
Table \ref{tab:gedit_benchmark} illustrates that HY-WU sets the new state-of-the-art for open-source models. In GEdit-Bench-EN, HY-WU achieves the top rank in semantic consistency and overall score, while securing the second-best perceptual quality. In GEdit-Bench-CN, our model achieves the best across all three metrics. Among closed-source models, HY-WU surpasses Seedream 4.5 and Nano-Banana-Pro across five out of six metrics and outperforms all models in the semantic consistency of GEdit-Bench-CN, highlighting HY-WU's strong capabilities.
\begin{table}[h!]
\centering
\tablestyle{8pt}{1.2}
\begin{tabular}{lc|ccc|ccc}
\multirow{2}{*}{model} &\multirow{2}{*}{public} & \multicolumn{3}{c|}{GEdit-Bench-EN} & \multicolumn{3}{c}{GEdit-Bench-CN} \\
& & Q\_SC $\uparrow$ & Q\_PQ $\uparrow$ & Q\_O $\uparrow$ 
& Q\_SC $\uparrow$ & Q\_PQ $\uparrow$ & Q\_O $\uparrow$ \\

\noalign{\vspace{1pt}}
\shline
\noalign{\vspace{1pt}}

Seedream 4.5 &\multirow{3}{*}{$\times$}
& \underline{7.920} & 7.427 & \underline{7.587}
& \underline{7.735} & 7.442 & 7.445 \\

GPT Image 1.5
&& \textbf{7.966} & \textbf{7.595} & \textbf{7.699} 
& \textbf{7.853} & \textbf{7.606} & \textbf{7.637} \\

Nano-Banana-Pro
&& 7.718 & \underline{7.476} & 7.391
& 7.725 & \underline{7.514} & \underline{7.470} \\

\noalign{\vspace{1pt}}
\shline
\noalign{\vspace{1pt}}

Step1X-Edit-v1.2& \multirow{4}{*}{$\checkmark$}
& 7.652 & 7.399 & 7.350 
& 7.491 & 7.433 & 7.265 \\

Qwen-Image-Edit-2511
&& \underline{7.898} & 7.380 & \underline{7.530}
& 7.723 & 7.335 & 7.378 \\

LongCat-Image-Edit
&& 7.832 & \textbf{7.503} & 7.519 
& \underline{7.820} & \underline{7.464} & \underline{7.516} \\

FLUX.2 [Dev]
&& 7.788 & 7.399 & 7.404 
& 7.593 & 7.344 & 7.247 \\

\rowcolor{gray!10} \textbf{HY-WU} 
& $\checkmark$ & \textbf{7.952} & \underline{7.496} & \textbf{7.593}
& \textbf{7.898} & \textbf{7.493} & \textbf{7.583}\\

\end{tabular}

\caption{Results on GEdit-Bench (Semantic
Consistency as Q\_SC, Perceptual Quality as Q\_PQ and Overall Performance as Q\_O). \textbf{Best} results are shown in bold and \underline{second-best} results are underlined.}
\label{tab:gedit_benchmark}
\end{table}

\paragraph{ImgEdit-Bench.}
We assess HY-WU's task-specific editing ability on ImgEdit-Bench~\citep{ye2025imgedit}. Table \ref{tab:imgedit_benchmark} shows that HY-WU achieves 5 top-1 and 1 top-2 results across the 9 sub-tasks among open-source models. With an overall score of 4.05, HY-WU ranks as the second-highest-performing public model and the third-highest overall, trailing the closed-source GPT Image 1.5 by a margin of 0.11 points.

\begin{table}[h!]
\centering
\tablestyle{3.9pt}{1.2}
\begin{tabular}{lc|ccccccccc|c} %
model & public & add & adjust & extract & replace & remove & background & style & hybrid & action & overall $\uparrow$ \\

\noalign{\vspace{1pt}}
\shline
\noalign{\vspace{1pt}}

Seedream 4.5
& \multirow{3}{*}{$\times$} & 4.63 & 3.92 & \underline{4.29} & 3.10 & 3.93 & 4.03 & \underline{3.20} & 3.25 & \underline{3.94} & 3.87 \\

GPT Image 1.5
& & \underline{4.77} & \textbf{4.49} & \textbf{4.44} & \textbf{3.54} & \textbf{4.37} & \textbf{4.37} & \textbf{3.30} & \underline{3.50} & \underline{3.94} & \textbf{4.16} \\

Nano Banana Pro
& & \textbf{4.81} & \underline{4.37} & 4.06 & \underline{3.41} & \underline{4.19} & \underline{4.29} & 3.19 & \textbf{3.55} & \textbf{4.11} & \underline{4.02} \\

\noalign{\vspace{1pt}}
\shline
\noalign{\vspace{1pt}}

Step1X-Edit-v1.2 
& \multirow{4}{*}{$\checkmark$} & 4.61 & \textbf{4.42} & 2.65 & 3.21 & 3.83 & 4.31 & 2.89 & 3.19 & 3.87 & 3.64 \\

Qwen-Image-Edit-2511
& & \underline{4.69} & 4.24 & 4.05 & 3.28 & 3.02 & 4.33 & 2.94 & \underline{3.50} & 3.80 & 3.77 \\

LongCat-Image-Edit
& & \textbf{4.72} & 4.22 & \textbf{4.44} & \underline{3.41} & \underline{4.31} & \underline{4.38} & \underline{3.14} & 3.19 & 3.99 & \textbf{4.07} \\

FLUX.2 [Dev] 
& & 4.64 & 3.90 & \underline{4.18} & \textbf{3.54} & 3.99 & 4.34 & 3.03 & 3.21 & \underline{4.01} & 3.92 \\

\rowcolor{gray!10} \textbf{HY-WU} 
& $\checkmark$ & 4.67 & \underline{4.38} & 4.12 & 3.37 & \textbf{4.34} & \textbf{4.45} & \textbf{3.28} & \textbf{3.56} & \textbf{4.02} & \underline{4.05} \\

\end{tabular}

\caption{Results on ImgEdit-Bench (Category-wise and Overall Performance). \textbf{Best} results are shown in bold and \underline{second-best} results are underlined.}
\label{tab:imgedit_benchmark}
\end{table}

\paragraph{WU-Eval.}
Table~\ref{tab:wueval} reports the quantitative comparison on the WU-Eval benchmark against both open-source and closed-source image editing models.
Among all evaluated methods, HY-WU achieves the best overall performance with an overall score of $4.27$, outperforming all public models and surpassing several leading closed-source systems.
In particular, HY-WU achieves the highest scores on consistency, structure, and quality, indicating stronger structural fidelity and more coherent visual edits.
Compared with strong open-source baselines such as Qwen-Image-Edit-2511 and FLUX.2, HY-WU consistently improves across all metrics, yielding gains of +0.27 in consistency and +0.23 in structure over the strongest open baseline.
These improvements highlight the effectiveness of instance-conditioned operator generation in maintaining edit stability and structural plausibility.
Notably, HY-WU also remains competitive with state-of-the-art closed-source models, achieving comparable or superior performance on multiple metrics.

\begin{table}[h!]
\centering
\tablestyle{5pt}{1.2}
\begin{tabular}{lc|ccccc}
model & public & alignment$\uparrow$ & consistency$\uparrow$ &structure$\uparrow$ &quality$\uparrow$ & overall$\uparrow$ \\

\noalign{\vspace{1pt}}
\shline
\noalign{\vspace{1pt}}

Seedream 4.5
& \multirow{3}{*}{$\times$} & 4.55 & \underline{4.02} & \textbf{4.24} & \textbf{3.89} & \underline{4.18} \\

GPT Image 1.5
& & \textbf{4.78} & 3.95 & 4.22 & 3.83 & \textbf{4.20} \\

Nano Banana Pro
& & \underline{4.58} & \textbf{4.02} & \underline{4.23} & \textbf{3.89} & \underline{4.18} \\

\noalign{\vspace{1pt}}
\shline
\noalign{\vspace{1pt}}

Step1X-Edit-v1.2 
& \multirow{4}{*}{$\checkmark$} & 3.83 & 3.52 & 3.72 & 3.35 & 3.61 \\

Qwen-Image-Edit-2511
& & \underline{4.54} & \underline{3.86} & \underline{4.07} & \underline{3.71} & \underline{4.04} \\

LongCat-Image-Edit
& & 4.47 & 3.83 & 4.03 & 3.66 & 4.00 \\

FLUX.2 [Dev] 
& & 4.37 & 3.81 & 4.01 & 3.68 & 3.97 \\

\rowcolor{gray!10} \textbf{HY-WU} 
& $\checkmark$ & \textbf{4.67} & \textbf{4.13} & \textbf{4.30} & \textbf{3.98} & \textbf{4.27} \\

\end{tabular}

\caption{Results on WU-Eval. \textbf{Best} results are shown in bold and \underline{second-best} results are underlined.}
\label{tab:wueval}
\end{table}

\subsection{Ablation Study}
\label{sec:ablation}
\begin{table*}[t]
\centering
\subfloat[\textbf{Foundation Model as Qwen-Image-Edit-2509}
\label{tab:foundation_model_qwen}]{
\centering
\begin{minipage}[h]{\textwidth}
\begin{center}
\centering
\tablestyle{4pt}{1.2}
\begin{tabular}{l|cccc|ccccc}
\multirow{2}{*}{method}& \multicolumn{4}{c|}{GSB (HumanEval)} 
& \multicolumn{5}{c}{WU-Eval}
\\
& G rate & S rate & B rate & win rate
& alignment$\uparrow$ & consistency$\uparrow$ &structure$\uparrow$ & quality$\uparrow$ & overall$\uparrow$
\\
\noalign{\vspace{1pt}}
\shline
\noalign{\vspace{1pt}}
\rowcolor{baselinecolor}
Qwen-Image-Edit-2509
&-&-&-&-
& 4.41 & 3.74 & 3.94 & 3.60 & 3.92 \\

\textbf{+ HY-WU}
& 17.7\% & 77.5\% & 4.8\% & 56.5\%
& 4.42 & 3.89 & 4.06 & 3.72 & 4.02 \\ %

\noalign{\vspace{1pt}}
\shline
\noalign{\vspace{1pt}}

improvement ($\uparrow$) 
& - & - & - & -
& +0.01 & +0.15 & +0.12 & +0.12 & +0.10 \\
\end{tabular}
\end{center}
\end{minipage}}
\vspace{5pt}

\subfloat[\textbf{Foundation Model as HY-Image-3.0-Instruct}
\label{tab:foundation_model_hunyuan}]{
\centering
\begin{minipage}[h]{\textwidth}
\begin{center}
\centering
\tablestyle{4pt}{1.2}
\begin{tabular}{l|cccc|ccccc}
\multirow{2}{*}{method}& \multicolumn{4}{c|}{GSB (HumanEval)} 
& \multicolumn{5}{c}{WU-Eval}
\\
& G rate & S rate & B rate & win rate
& alignment$\uparrow$ & consistency$\uparrow$ &structure$\uparrow$ &quality$\uparrow$ & overall$\uparrow$
\\
\noalign{\vspace{1pt}}
\shline
\noalign{\vspace{1pt}}
\rowcolor{baselinecolor}
HY-Image-3.0-Instruct
&-&-&-&-
& 4.69 & 4.03 & 4.26 & 3.96 & 4.23\\

\textbf{+ HY-WU}
& 9.5\% & 87.0\% & 3.5\% & 53.0\%
& 4.67 & 4.13 & 4.30 & 3.98 & 4.27 \\

\noalign{\vspace{1pt}}
\shline
\noalign{\vspace{1pt}}

improvement ($\uparrow$) 
& - & - & - & -
& -0.02 & +0.10 & +0.04 & +0.02 & +0.04 \\
\end{tabular}
\end{center}
\end{minipage}}

\vspace{5pt}

\subfloat[\textbf{Scaling Up LoRA Parameter}
\label{tab:scaling_lora}]{
\centering
\begin{minipage}[h]{0.45\textwidth}
\begin{center}
\centering
\tablestyle{7pt}{1.2}
\begin{tabular}{l|c}
method & win rate
\\

\noalign{\vspace{1pt}}
\shline
\noalign{\vspace{1pt}}

\rowcolor{baselinecolor}
Qwen-Image-Edit-2509
&-\\

\noalign{\vspace{1pt}}
\shline
\noalign{\vspace{1pt}}

\textbf{+ HY-WU} w/ LoRA (rank=16, \textbf{0.12B})
&51.4\%\\

\textbf{+ HY-WU} w/ LoRA (rank=32, \textbf{0.24B})
&51.0\%\\

\textbf{+ HY-WU} w/ LoRA (rank=64, \textbf{0.47B})
&53.7\%\\

\end{tabular}
\end{center}
\end{minipage}}
\subfloat[\textbf{Scaling Up NNT Parameter}
\label{tab:scaling_nnt}]{
\centering
\begin{minipage}[h]{0.45\textwidth}
\begin{center}
\centering
\tablestyle{7pt}{1.2}
\begin{tabular}{l|c}
method & win rate
\\

\noalign{\vspace{1pt}}
\shline
\noalign{\vspace{1pt}}

\rowcolor{baselinecolor}
Qwen-Image-Edit-2509
&-\\

\noalign{\vspace{1pt}}
\shline
\noalign{\vspace{1pt}}

\textbf{+ HY-WU} w/ \textbf{2B} parameter NNT
&51.0\%\\

\textbf{+ HY-WU} w/ \textbf{5B} parameter NNT
&53.8\%\\

\textbf{+ HY-WU} w/ \textbf{7B} parameter NNT
&53.8\%\\

\end{tabular}
\end{center}
\end{minipage}}

\caption{\textbf{Ablation studies about foundation model and scaling behavior. (Foundation model is marked with gray stripe.)} For foundation model ablation (a) and (b), HY-WU achieve consistent improvement across different model architecture, highlighting its universality. For scaling behavior ablation (c) and (d), results demonstrate a consistent positive scaling law across both LoRA parameter and Neural Network Transformer (NNT) parameter.}
\label{tab:ablations}

\end{table*}

\paragraph{Foundation Model Backbones.}
To assess the architectural universality of the HY-WU framework, we examine it with two distinct backbone architectures: (1) a transfusion-based~\citep{Zhou2024Transfusion} native unified multimodal model, represented by HY-Image-3.0-Instruct.
(2) the traditional multimodal diffusion transformer (MMDiT)~\citep{Esser2024ScalingRF}, represented by Qwen-Image-Edit-2509.
For each backbone, we compare the original foundation model with its HY-WU-integrated variant using both human evaluation (GSB) and automatic evaluation (WU-Eval).
As shown in Table \ref{tab:ablations}, HY-WU consistently improves the overall WU-Eval score across both architectures, with gains of $+0.10$ and $+0.04$ respectively.
The improvements are most pronounced in \textit{consistency} and \textit{structure}, while alignment remains comparable.
This pattern aligns with the design goal of HY-WU: improving structural fidelity and edit consistency while preserving instruction alignment.

The GSB human evaluation further confirms these improvements.
HY-WU achieves win rates of 56.5\% and 53.0\% against the corresponding base models, indicating that the improvements are perceptually meaningful rather than metric-specific artifacts.
Together, these results demonstrate that HY-WU generalizes across heterogeneous generative architectures.

\paragraph{Scaling Law.}
We further analyze the scalability of HY-WU along two dimensions: 
(1) the capacity of the neural network transformer (NNT) used for parameter generation, and 
(2) the capacity of the generated LoRA parameters.
For transformer scaling, we vary the generator size to 2B, 5B, and 7B parameters by increasing the transformer's depth.
For LoRA scaling, we increase the LoRA rank to obtain effective parameter sizes of 0.12B (rank 16), 0.24B (rank 32), and 0.47B (rank 64).
As shown in Table~\ref{tab:ablations}, both scaling trajectories generally exhibit improved performance with larger capacity.
Increasing the NNT size leads to a consistent improvement in win rate, reaching 53.8\% for both the 5B and 7B configurations.
Similarly, larger LoRA parameter budgets tend to improve editing performance, with the highest win rate achieved at rank=64.
These results suggest that HY-WU benefits from increased representational capacity in both the generator and the generated adapters, indicating favorable scaling behavior.
Note that results are obtained in a small-scale experimental setting. Thus, performance may vary from other experiments.

\subsection{Qualitative Results}
\label{sec:qualitative}

\begin{figure*}[t!]
    \centering
    \vspace{-1em}
    \includegraphics[width=1.0\linewidth]{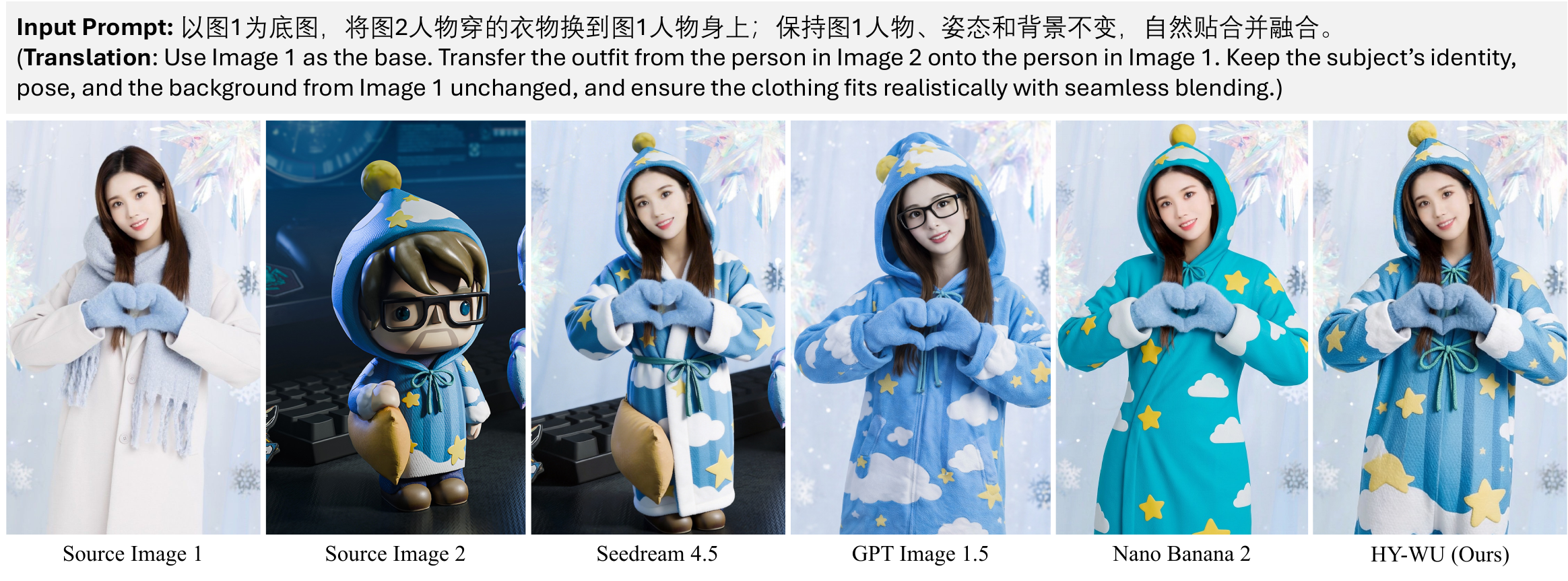}
    \includegraphics[width=1.0\linewidth]{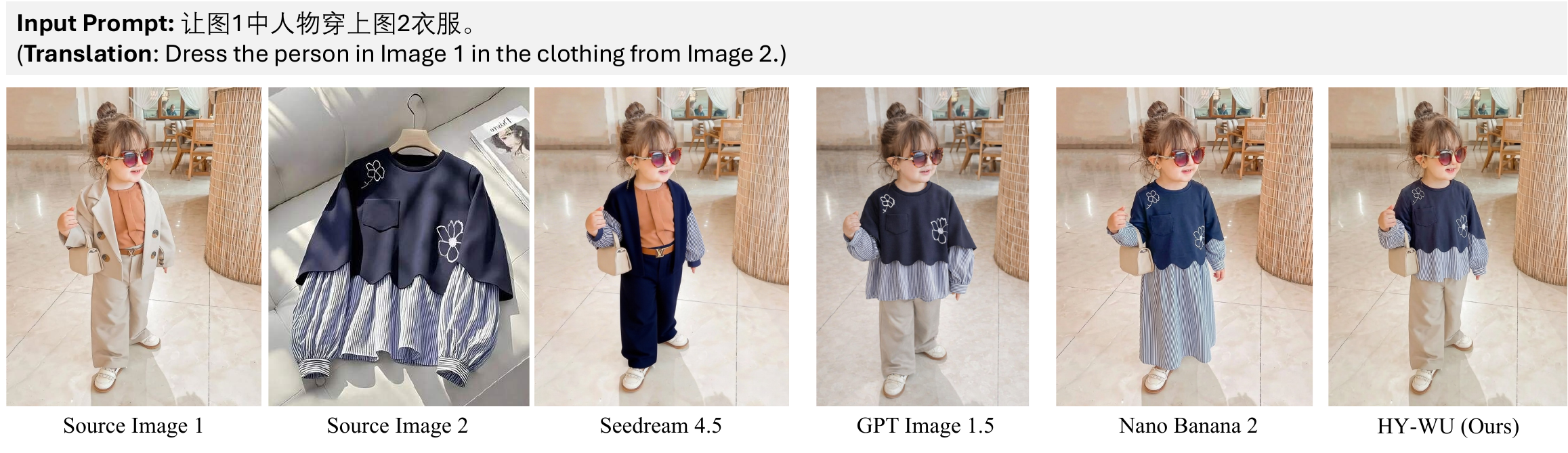}
    \includegraphics[width=1.0\linewidth]{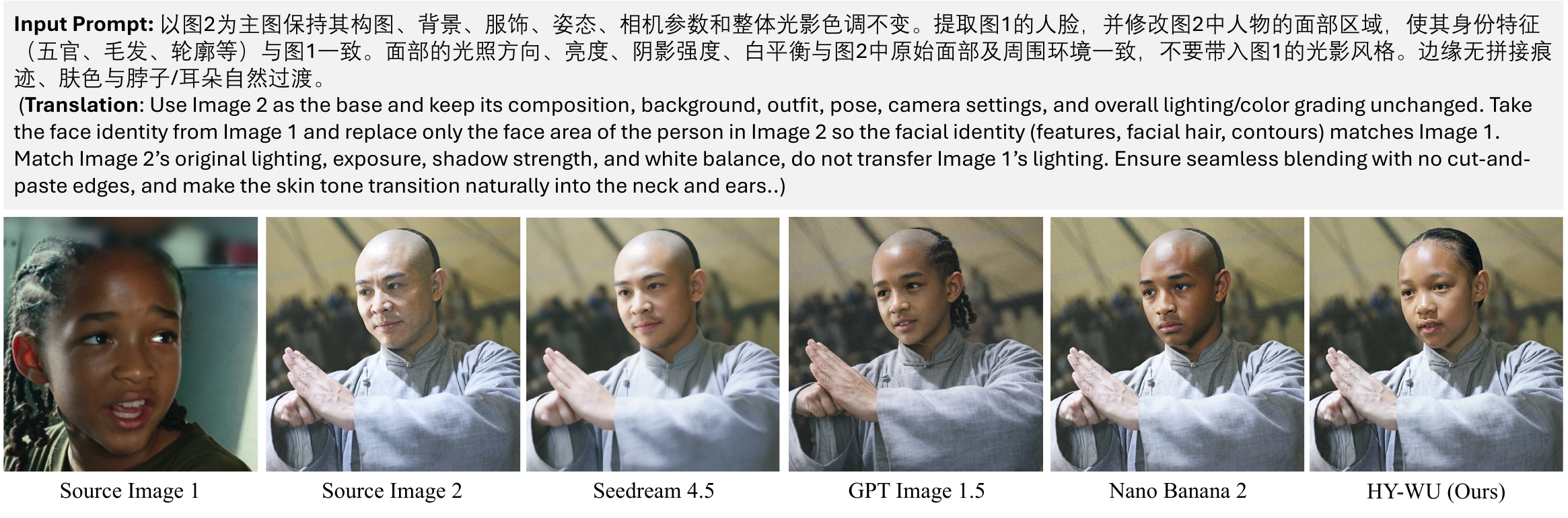}
    \vspace{-2em}
    \caption{\textbf{Visual comparison on multi-image editing tasks under consistency constraints.}
    Each task requires transferring attributes from a reference image while preserving key invariants of the base image (e.g., identity, pose, background, or lighting).
    Competing methods often introduce inconsistencies such as identity drift, structural distortion, or imperfect blending.
    In contrast, HY-WU produces edits that better preserve subject identity, pose, and scene context while faithfully transferring the target attributes.
    }
    \label{fig:comp_gaming}
    \vspace{-1.3em}
\end{figure*}

To further demonstrate the visual editing capacity of HY-WU, we provide a qualitative comparison against state-of-the-art closed-source models (Seedream 4.5, GPT Image 1.5, and Nano Banana 2).
Figure~\ref{fig:comp_gaming} presents qualitative comparisons on multi-image editing tasks, including clothing transfer and identity replacement.
These tasks require simultaneously preserving the base image identity, pose, and scene context while integrating attributes from a reference image.

As shown in Figure~\ref{fig:comp_gaming}, several baselines introduce identity drift, structural distortion in clothing geometry, or unnatural blending between transferred attributes and the original subject.
In contrast, HY-WU produces edits that remain more faithful to the base image while accurately transferring the reference attributes.
Clothing patterns and geometry are better aligned with the body pose, and facial edits preserve natural lighting and structure.
This improved consistency suggests that instance-conditioned operator generation enables more reliable execution of editing rules, avoiding the compromise behavior often observed in static adaptation methods.

We provide additional qualitative showcases in Figure~\ref{fig:showcase_gaming1},\ref{fig:showcase_gaming2},\ref{fig:showcase_doll},\ref{fig:showcase_tryon},\ref{fig:showcase_faceswap}.
These examples cover a diverse set of multi-image editing scenarios, including character costume transfer in game posters, toy-to-human clothing transfer, virtual try-on, and identity-preserving face replacement.
Across these tasks, HY-WU consistently preserves key visual invariants such as subject identity, pose, and background while accurately transferring attributes from the reference image.
Notably, the generated edits remain structurally coherent even when the source and reference images differ significantly in style, semantics, or domain (\eg, stylized game characters \textit{vs.} real photographs, or doll costumes \textit{vs.} human garments).
These results highlight the ability of HY-WU to generalize across various editing scenarios while maintaining strong visual consistency.

\section{Analyses and Discussions}
In this section, we investigate the structural behavior of static adaptation and the geometric properties of conditional parameter generation.
The main performance comparisons in Section~\ref{sec:main_performance} are performed on HY-Image-3.0-Instruct~\citep{cao2025hunyuanimage} backbone, a native unified multimodal model, to demonstrate state-of-the-art capability.
In this section, all analyses are conducted on the Qwen-Image-Edit-2509~\citep{wu2025qwen} backbone, a pure diffusion model, for controlled and reproducible investigation of adaptation dynamics within a standard diffusion operator.
Unless otherwise stated, the conclusions in this section concern the adaptation mechanism (static \textit{vs.} conditional) rather than backbone-specific characteristics.
Throughout this section, we refer to the core mechanism of HY-WU as \textbf{Parameter Generation (PG)}, emphasizing the difference between static and conditional operators rather than the specific implementation details of the full framework.

Here are key takeaways according to our findings:

\begin{tcolorbox}[colback=gray!8, colframe=black!60, title=\textbf{Takeaways}]
\begin{itemize}
    \item \textbf{Static sharing compromises under conflict:} When editing objectives are mutually exclusive, shared static adaptation converges to compromise operators, visible both behaviorally (conflict-controlled edits) and mechanistically (gradient-level conflict) (Section~\ref{sec:behavioral_conflict} and Section~\ref{sec:grad_conflict}). Conditional parameter generation avoids this by routing instances to a family of operator updates.
    \item \textbf{Routing, not capacity, is decisive:} Instance-level gains require correct condition--parameter alignment. Destroying alignment (Average PG / Shuffle PG) collapses performance toward base-level despite similar parameter count, isolating conditional routing as the key mechanism (Section~\ref{sec:instance_adaptivity}).
    \item \textbf{Emergent semantic structure in update space:} Generated updates form a semantically organized parameter manifold without explicit parameter-space supervision. Both global clusters and local kNN neighborhoods in parameter space align with image and text semantics (Section~\ref{sec:parameter_structure}).
    \item \textbf{Static updates collapse into narrow regions:} Directly optimized static adapters concentrate near initialization and overlap across tasks, whereas conditional generation explores a broader structured region consistent with a conditional update family (Section~\ref{sec:train_vs_pg}).
\end{itemize}
\end{tcolorbox}

Together, these findings support a reformulation of adaptation: instead of optimizing a single shared parameter point, effective adaptation should learn a conditional family of parameter space.

\subsection{Analysis of Infeasible Shared Optimization}
\label{sec:behavioral_conflict}
\begin{figure*}[t]
    \centering
    \begin{subfigure}[t]{1\textwidth}
        \centering
        \includegraphics[width=\textwidth]{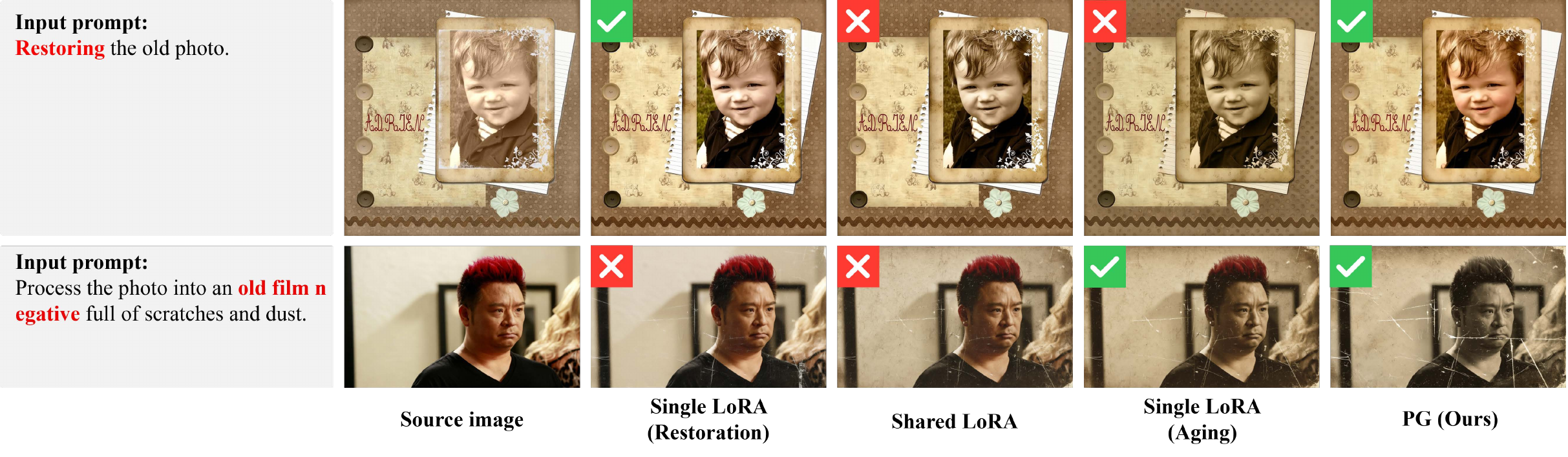}
        \caption{Restoration \textit{vs.} Aging.}
        \label{fig:restore_vs_age}
    \end{subfigure}
    \hfill
    \begin{subfigure}[t]{1\textwidth}
        \centering
        \includegraphics[width=\textwidth]{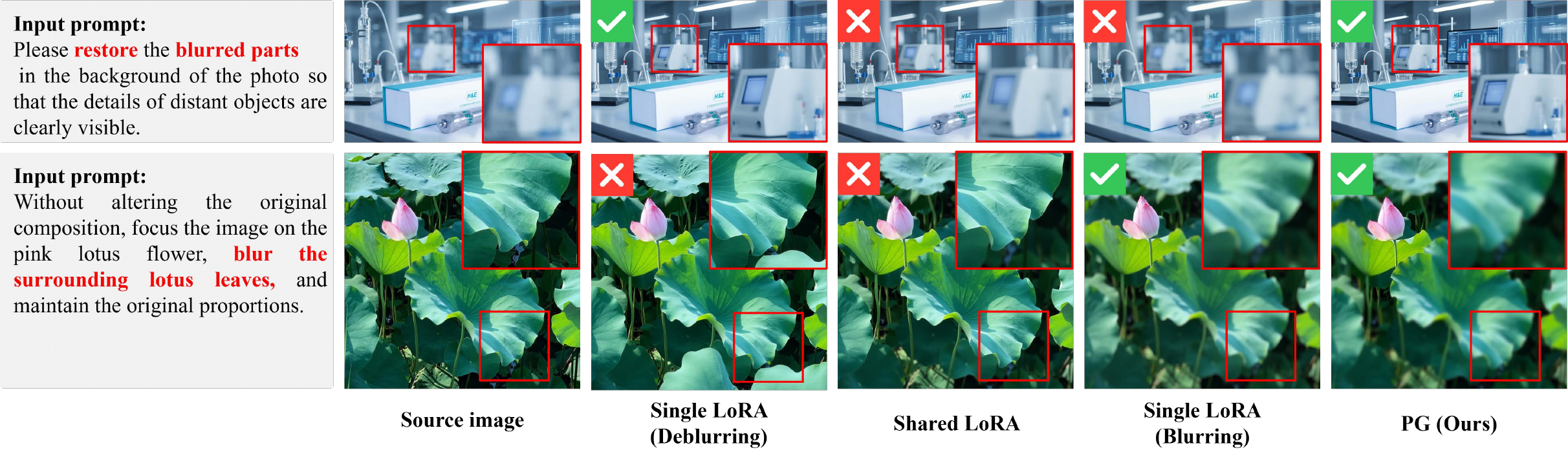}
        \caption{Deblurring \textit{vs.} Blurring.}
        \label{fig:deblur_vs_blur}
    \end{subfigure}
    \vspace{-3mm}
    \caption{\textbf{Behavioral evidence of structural compromise under shared static adaptation in conflict settings.}
    For each conflicting task pair, we compare:
    (1) Single LoRA trained per objective,
    (2) Shared LoRA trained jointly on both objectives,
    and (3) PG trained jointly.
    Green check marks indicate correct directional behavior, and red crosses indicate failure.
    \textbf{(a) Restoration vs. Aging.}
    Restoration enhances clarity and recovers natural color, whereas aging introduces degradation and discoloration.
    Single LoRA models exhibit strong directional behavior when trained separately, but they fail on the opposing task.
    Shared LoRA produces intermediate outputs: restored images remain slightly desaturated, and aged outputs appear insufficiently degraded.
    PG performs well for both objectives.
    \textbf{(b) Deblurring vs. Blurring.}
    Deblurring restores high-frequency details, while blurring suppresses them.
    Shared LoRA yields softened results in both directions.
    In contrast, PG maintains sharper deblurring and better blurring effects.
    Across both settings, single LoRA models exhibit over-specialization behavior and fail on the opposing task. Shared adaptation converges to a compromise operator.
    PG maintains clearer objective-specific behavior across both objectives, suggesting that condition-dependent parameters reduce the need for compromise within shared parameter space.
    }
    \label{fig:shared_lora_conflict}
\end{figure*}

To investigate whether static adaptation (LoRA / SFT) suffers from infeasible optimization under mutually exclusive objectives, we carefully construct controlled conflict settings using two opposing task pairs:
(1) restoration \textit{vs.} intentional aging, and 
(2) deblurring \textit{vs.} blurring.
These task pairs are structurally contradictory.
Restoration and deblurring enhance high-frequency detail and recover natural structure,
while aging and blurring intentionally suppress or degrade detail.
As these objectives are inherently directional and visually contradictory, optimizing both objectives within a single shared parameter update imposes competing directional constraints.

For each task pair, we compare three adaptation strategies under identical training budgets:
\textbf{Single LoRA} (separate adapter trained per objective),
\textbf{Shared LoRA} (a single static adapter trained jointly on both objectives),
and \textbf{PG} (instance-conditioned adapter generation trained jointly).
Note that each objective contains 1k samples for training.
Evaluation is conducted on a held-out subset, with qualitative inspection focusing on cases where directional effects are clearly demonstrated.
Representative examples are shown in Figure~\ref{fig:shared_lora_conflict}. We observe the following consistent patterns for both conflict settings:

\textbf{(i) Single LoRA.}
When trained on a single objective, the model exhibits strong and coherent directional transformations toward training objective. However, this over-specialization prevents it from performing well on the opposing task.

\textbf{(ii) Shared LoRA.}
When trained jointly on opposing objectives, the static adapter partially satisfies both tasks but tends to produce visually compromised outputs in shared settings.
In restoration \textit{vs.} aging (Figure~\ref{fig:restore_vs_age}), restored images from shared LoRA remain slightly desaturated or under-enhanced, while aging outputs appear insufficiently degraded.
In deblurring \textit{vs.} blurring (Figure~\ref{fig:deblur_vs_blur}), shared LoRA cannot recover sharp details from blurry screen, while bottom outputs exhibit compromised results in blurring the lotus leaves.
These patterns indicate that the shared parameter update balances conflicting gradients rather than committing to a clear direction.

\textbf{(iii) Parameter Generation.}
PG maintains clearer directional behavior across both objectives.
Restoration and deblurring outputs remain sharp and well-defined, while results on aging and blurring samples are more consistent with the intended objective.
This suggests that condition-dependent parameter generation reduces the need to embed incompatible update directions within a single static update.

These findings provide behavioral evidence of infeasible shared optimization. When heterogeneous samples impose mutually incompatible constraints on a shared update, static adaptation converges toward a compromise solution.
In contrast, conditional parameter generation allocates condition-dependent parameters, preserving directional consistency without collapsing into a single shared compromise.

\subsection{Analysis of Gradient-Level Conflict}
\label{sec:grad_conflict}
\begin{figure*}[t]
    \centering
    \includegraphics[width=\linewidth]{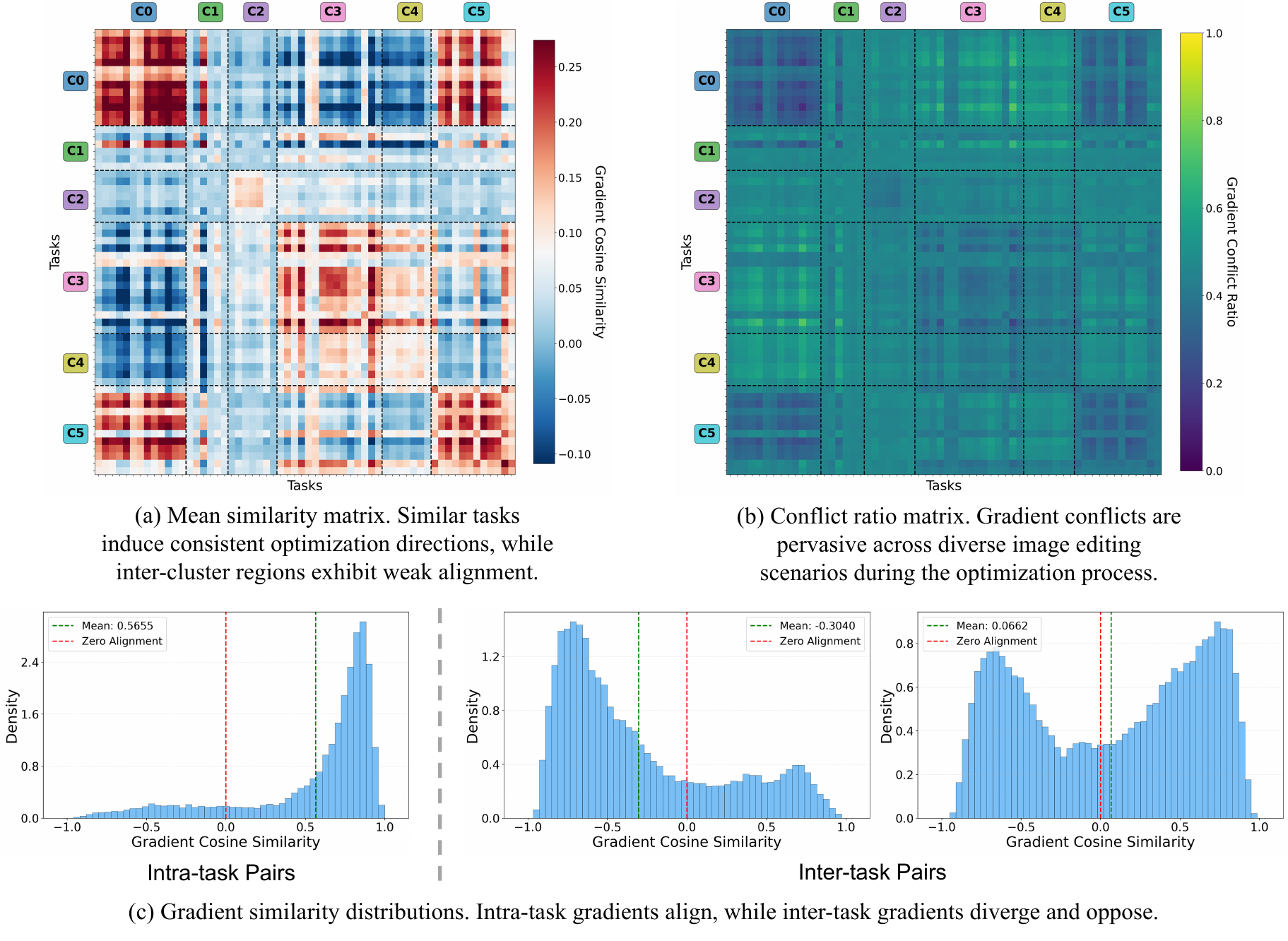}
    \caption{\textbf{Gradient-Level Conflict Structure across Editing Tasks.}
    \textbf{(a)} Mean cosine similarity among tasks. We cluster similar tasks into C0 - C5 for visualization clarity. Each node represents a mean of gradient similarity among samples from two tasks.
    \textbf{(b)} Conflict ratio matrix, defined as the proportion of negative cosine values. Lighter color indicates higher conflict ratio.
    \textbf{(c)} Representative histogram distributions illustrating different modes, including intra-task pair (left) and inter-task pairs (middle and right). These distributions demonstrate clear pattern, \eg, strong positive bias, negative bias, and bimodality alignment.
    Together, these statistics reveal that heterogeneous editing objectives impose structured and often incompatible optimization directions under shared parameters.
    }
    \label{fig:grad_conflict}
\end{figure*}

Beyond the behavioral analysis of interference in shared optimization in Section~\ref{sec:behavioral_conflict}, we further directly analyze gradient conflict in optimization.
Specifically, we examine whether heterogeneous image-editing objectives produce structured gradient interference under a shared LoRA parameterization.
We examine a LoRA module on 60 editing tasks using 12k samples (200 samples per task). 
For each sample, we compute gradients of the loss with respect to the LoRA parameters and then measure gradient cosine similarity from:
(i) different samples within the same task (intra-task), and 
(ii) samples from different tasks (inter-task).
To characterize global structure, we construct two task-level statistics:
(1) the \emph{mean cosine similarity} between task pairs, and 
(2) the \emph{conflict ratio}, defined as the proportion of negative cosine values between task pairs.
Similarity matrices are reordered and clustered (C0 - C5) via spectral clustering for visualization clarity (Figure~\ref{fig:grad_conflict}).
Here, clustering is performed at the \emph{task level}, \ie, each node in the matrix corresponds to an editing task pair, and clusters group tasks that exhibit higher aligned gradients.

\paragraph{Global conflict structure.}
Figure~\ref{fig:grad_conflict}(a) shows the mean cosine similarity between task pairs.
Red regions indicate positive alignment (gradients from different samples are pointing in similar directions), while blue regions indicate negative alignment (opposite directions).
Strong red blocks appear along intra-task (diagonal) regions, indicating that samples from the same task induce consistent optimization directions.
In contrast, many cross-task regions exhibit weak alignment (near white) or clear negative values (blue), suggesting directional disagreement between heterogeneous editing objectives.
Figure~\ref{fig:grad_conflict}(b) reports the conflict ratio matrix.
Brighter colors indicate a higher proportion of negative gradient similarities among samples.
Several task pairs exhibit high conflict ratios, meaning that a significant proportion of gradient components are directionally opposed, leading to opposite optimization directions.
Notably, both matrices reveal consistent patterns: gradient conflict exists prevalently in optimization.

\paragraph{Distributional patterns.}
Figure~\ref{fig:grad_conflict}(c) illustrates several representative gradient similarity distributions.
The left histogram (intra-task) concentrates strongly on the positive side with a mean $\approx 0.56$, indicating coherent task-specific gradients.
The middle example  (inter-task) shows a predominantly negative distribution (mean $\approx -0.30$), reflecting fundamentally opposed optimization directions between two tasks.
The right example exhibits another inter-task pair with a bi-modal shape, where substantial mass distribute on both positive and negative regions.
Such bimodality indicates co-existence of shared and conflicting gradient subspaces.
These observations demonstrate that gradient conflict across editing tasks is structured and pervasive rather than random noise.

Under static training, gradients are aggregated into a single update direction.
When a substantial fraction of gradient components are misaligned or opposed, aggregation inevitably produces a compromise direction that cannot simultaneously optimize all objectives.
The observed gradient-level interference therefore provides mechanistic evidence that heterogeneous objectives cannot be simultaneously satisfied by a single shared parameter point.

\subsection{Instance-Level Adaptivity}
\label{sec:instance_adaptivity}
\begin{table*}[t]

\centering
\tablestyle{8pt}{1.2}
\begin{tabular}{l|c|c|ccccc}
\textbf{model}  & \textbf{conditional} & \textbf{win rate $\uparrow$} & \textbf{alignment $\uparrow$} & \textbf{consistency $\uparrow$} & \textbf{structure $\uparrow$} & \textbf{quality $\uparrow$} & \textbf{overall $\uparrow$} \\

\noalign{\vspace{1pt}}
\shline
\noalign{\vspace{1pt}}
(a) base model      &    \cmark      & - & 4.41      & 3.74        & 3.94      & 3.60    & 3.92    \\
(b) shared LoRA     &    \xmark    & 51.7\% & 4.41      & 3.83        & 3.99      & 3.67    & 3.98    \\
(c) SFT             &    \xmark    & 51.5\% & \textbf{4.45}      & 3.84        & 4.02      & 3.65    & 3.99    \\
(d) average PG      &    \xmark    & 48.0\% & 4.40      & 3.74        & 3.94      & 3.62    & 3.92    \\
(e) shuffle PG      &    \cmark (misaligned) & 48.3\% & 4.38      & 3.76        & 3.97      & 3.62    & 3.93    \\
(f) full PG         &   \cmark     & 56.5\% & 4.42      & \textbf{3.89}       & \textbf{4.06}      & \textbf{3.72}    & \textbf{4.02}   \\
\end{tabular}

\caption{\textbf{Quantitative evaluation of instance-conditioned adaptivity.}
PG achieves the highest overall performance and dominates in consistency, structure, and quality, while alignment remains comparable across methods. 
SFT and Shared LoRA, though differing significantly in parameter capacity, yield similar overall performance.
Average PG (using averaged generated LoRA across samples as a fixed parameter) collapses to the base-model level, and Shuffle PG (randomly permuting conditional inputs to PG during inference) degrades toward non-conditional baselines, indicating that gains come from input-aligned instance-specific parameter updates rather than architectural capacity.}
\label{tab:instance_ada}
\end{table*}

We investigate the core mechanism behind PG: whether its advantage arises from instance-conditioned parameter generation or merely from increased parameter capacity relative to static adaptation.
We compare six variants: 
(a) Base model,
(b) Shared LoRA,
(c) SFT (full supervised fine-tuning),
(d) Average PG (compute the mean LoRA parameter across a large sample set and use this fixed parameter for all inputs during inference). 
(e) Shuffle PG (randomly shuffling the PG input conditions across samples during inference).
(f) Full PG.
We evaluate all variants using both automatic WU-Eval metrics and human GSB win rate (pairwise preference against the base model), providing complementary assessments.

Table~\ref{tab:instance_ada} shows that full PG achieves the highest win rate (56.5\%) and the best overall WU-Eval score (4.02).
The improvements are consistent across Consistency, Structure, and Quality, while Alignment remains comparable to static baselines.
This alignment between human preference and automatic metrics indicates that the gains are perceptually meaningful rather than metric-specific artifacts.
These improvements indicate that instance-conditioned parameter generation primarily enhances controllability and structural fidelity, reducing undesired changes in non-edited regions, improving structural plausibility, and suppressing artifacts.

Interestingly, full supervised fine-tuning (SFT) and Shared LoRA obtain nearly identical win rates (51.5\% vs 51.7\%) and similar overall WU-Eval scores (3.99 vs 3.98).
Despite its substantially larger parameter capacity, SFT remains a static optimization of a single shared parameter point. 
This suggests that increased capacity alone within a static shared update does not eliminate the structural limitation of single-point adaptation.
In contrast, PG consistently improves structural and perceptual metrics beyond both static baselines.

Crucially, when instance–parameter correspondence is removed,
Average PG and Shuffle PG collapse toward base-level performance in both win rate (48.0\%, 48.3\%) and WU-Eval metrics.
Since these variants preserve parameter count and compute budget but destroy conditional alignment, the degradation isolates routing as the decisive factor.
This demonstrates that correct instance–parameter alignment is necessary: misaligned LoRA parameters greatly harm the performance.

Together, these results indicate that PG’s advantage does not stem from additional parameters, but from dynamically assigning instance-specific operator updates.
Removing this alignment collapses PG into a static approximation, eliminating its benefits.
These findings support that effective adaptation should be formulated as learning a conditional family of parameter space, rather than optimizing a single shared parameter point.

\subsection{Emergent Structure in Generated Parameter Space}
\label{sec:parameter_structure}

To examine whether conditional parameter generation learns structured parameter behavior rather than producing arbitrary mappings, we analyze the geometry of generated LoRA weights at the instance level.
In particular, we collect 12k editing samples covering a broad spectrum of domains, scenes, objects, and editing operations. 
For each input, we generate its LoRA parameters using a pretrained PG model. Each LoRA weight is flattened and projected to a 5,760-dimensional space via random projection to reduce computational cost, which approximately preserves pairwise Euclidean distances (Johnson–Lindenstrauss Lemma~\citep{JohnsonLindenstrauss1984}).

\begin{figure*}[t]
    \centering
    \includegraphics[width=\linewidth]{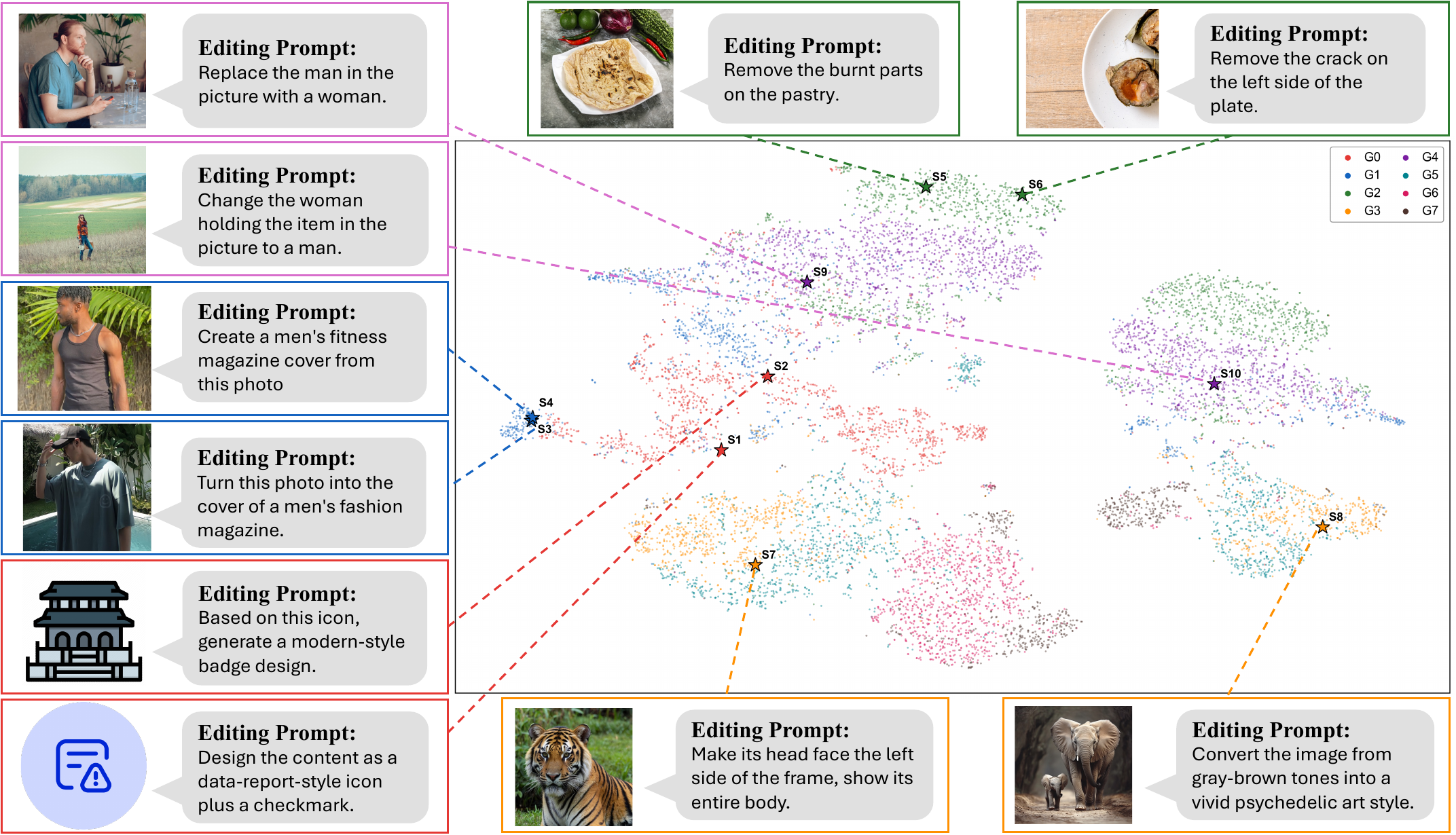}
    \caption{\textbf{Emergent semantic structure in parameter space learned by PG.}
    Samples are clustered with K-means in multimodal semantic space (image + text embeddings) and color-coded and visualized in LoRA parameter space via t-SNE. 
    Semantically related edits form coherent regions in parameter space, with representative examples highlighted. 
    Note that clusters may consist of multiple sub-regions, indicating a hierarchical structure, yet remain globally distinct across editing semantics.
    This suggests that conditional parameter generation induces a structured parameter landscape rather than collapsing into a single shared adapter.
}
    \label{fig:tsne}
\end{figure*}

\paragraph{Global Clustering Analysis.}
We first investigate whether semantic structure in editing inputs is reflected in the global geometry of generated parameters.
We construct a semantic embedding for each sample by concatenating its CLIP image and BGE text embeddings extracted from BAAI/bge-large-zh-v1.5~\citep{xiao2024c}.
These multimodal representations are projected to a lower-dimensional space and clustered using K-Means ($K=8$) in an unsupervised manner.
The resulting cluster assignments (G0 - G7), derived purely from data semantics, are then used to color-code the corresponding LoRA embeddings in parameter space, which are visualized via t-SNE for qualitative inspection.
Note that this clustering is performed at the \emph{sample level}, \ie, each point corresponds to an individual image–instruction pair.
This is distinct from the \emph{task-level} spectral clustering used in Section~\ref{sec:grad_conflict}, where clusters group entire tasks based on gradient interaction patterns.

As shown in Figure~\ref{fig:tsne}, semantically clustered samples form well-separated regions in LoRA parameter space. 
Editing types such as pose modification, content removal, logo modification, and style transformation naturally organize into distinct parameter regions.
Importantly, clustering is performed without access to LoRA weights during assignment, indicating that semantic structure in data space aligns with geometric structure in generated parameter space.

Representative samples (highlighted with star markers) further illustrate this alignment. 
For instance, samples related to removing objects ({\color{green} S5} and {\color{green} S6}) concentrate within a compact region characterized by similar images and editing prompts.
Logo modifications ({\color{red} S1} and {\color{red} S2}) form another cluster, reflecting their shared structural constraints, and edits that involve magazine designs from humans ({\color{blue} S3} and {\color{blue} S4}) tend to group together.
Interestingly, some semantic clusters appear as multiple nearby sub-regions rather than a single compact blob in parameter space. 
For example, {\color{orange} S7} and {\color{orange} S8} both involve animal-related edits, yet they diverge in editing tasks (\ie, pose change \textit{vs} style change).
A complementary pattern appears in {\color{purple} S9} and {\color{purple} S10}, which share the same editing operation, \ie, human gender swapping, but differ in source image semantics (man \textit{vs} woman). 
Despite such local fragmentation, these sub-regions remain globally coherent and clearly separated from unrelated semantic clusters. 
This pattern suggests that the generated parameter space captures hierarchical organization: high-level editing semantics define coarse cluster structure, while finer-grained variations in object identity, transformation direction, or scene context further structure parameters within each cluster.

Therefore, the generated parameter space is not rigidly partitioned by discrete task labels or text prompts, but organized as a structured and multi-modal manifold reflecting continuous semantic variation.
This pattern suggests that the learned parameter space exhibits hierarchical organization, \ie, high-level editing semantics define coarse clusters, while finer-grained factors like object identity and scene semantics further structure parameters within each sub-region.
This observation supports that the learned parameter space is structured but smoothly organized, reflecting continous semantic variation rather than hard categorical boundaries.

\begin{figure*}[t]
    \centering
    \begin{subfigure}[t]{0.48\textwidth}
        \centering
        \includegraphics[width=\textwidth]{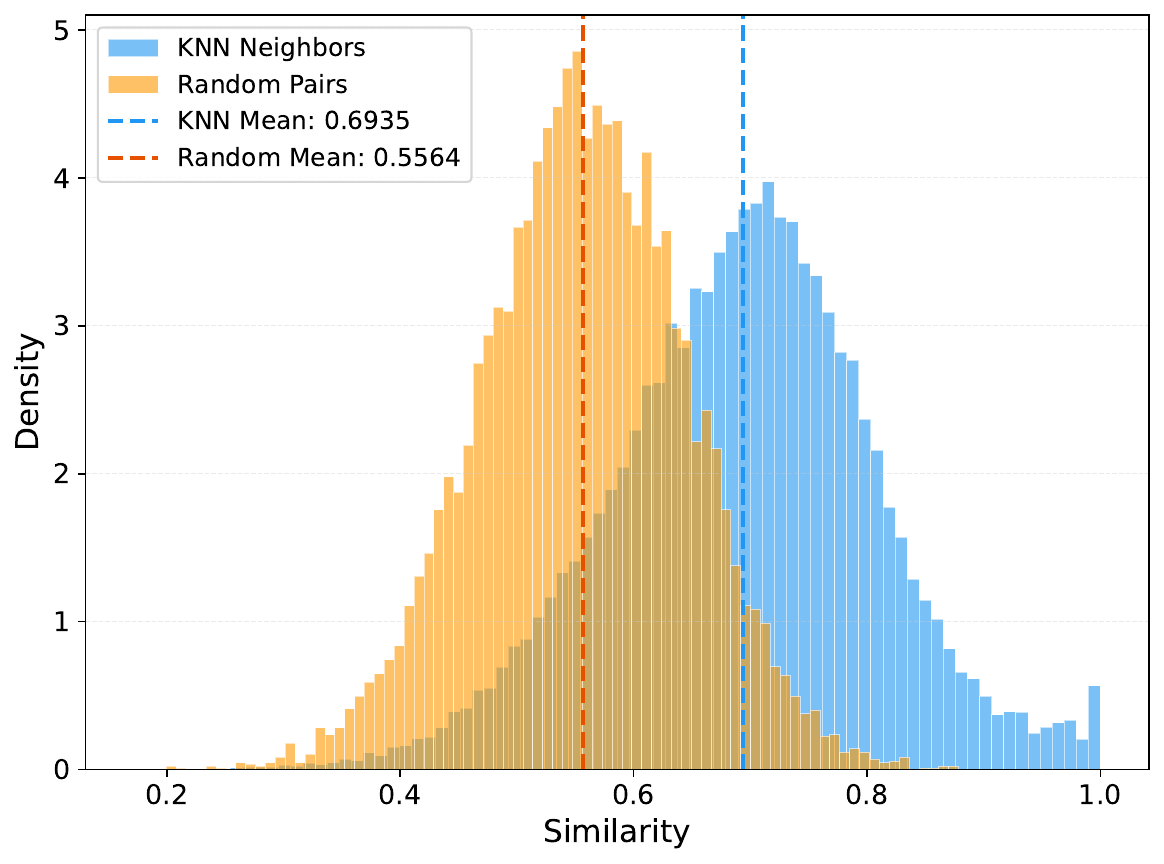}
        \caption{Image embedding similarity distribution.}
        \label{fig:hist_img_sim}
    \end{subfigure}
    \hfill
    \begin{subfigure}[t]{0.48\textwidth}
        \centering
        \includegraphics[width=\textwidth]{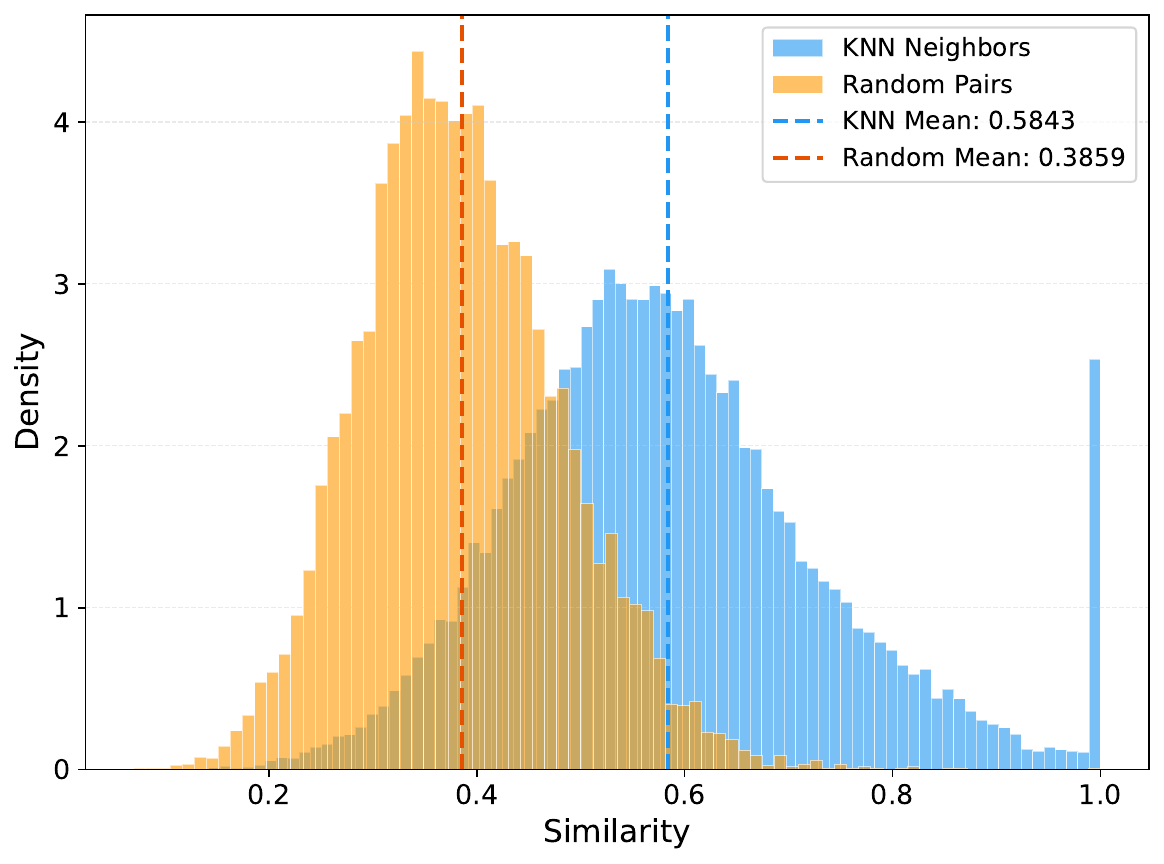}
        \caption{Text embedding similarity distribution.}
        \label{fig:hist_txt_sim}
    \end{subfigure}
    \caption{\textbf{Local semantic consistency induced by PG (kNN in LoRA parameter space)}.
    Neighbor pairs defined in parameter space consistently exhibit higher semantic similarity than random pairs in both image and text modalities, indicating that proximity in the generated parameter space aligns with semantic proximity, providing quantitative evidence of structured, condition-dependent organization in parameter space.
    We report distributions of \textbf{(a) image embedding similarity} and \textbf{(b) text embedding similarity}, compared against randomly sampled pairs.
    For each sample, we retrieve its five nearest neighbors in the projected LoRA parameter space and compute pairwise semantic similarity between neighbor pairs with editing instructions and input images. 
    }
    \label{fig:hist_knn}
\end{figure*}

\paragraph{Local Semantics Alignment Analysis.}
To quantitatively investigate local structure, we analyze semantic consistency within parameter-space neighborhoods. For each sample, we retrieve its $k$ nearest neighbors ($k=5$) in the projected LoRA parameter space. 
Given a sample $(x_i, y_i)$ with image $x_i$ and instruction editing prompt $y_i$, we compute semantic similarity for all neighbor pairs using:
(i) CLIP \citep{radford2021learning} image embedding similarity. 
(ii) Text embedding similarity.
For textual semantic similarity, we adopt BAAI/bge-large-zh-v1.5 \citep{xiao2024c} instead of CLIP text embeddings. 
We observe that CLIP text embeddings exhibit strong saturation for editing prompts, with many prompts clustering in a high-similarity regime (>0.9), limiting discriminative power for fine-grained editing semantics. 
In contrast, BGE-large-zh-v1.5, a bilingual retrieval-oriented embedding model, provides better semantic resolution for instruction-level differences. 
As a baseline, we also compute the same similarities for randomly sampled pairs.
As shown in Figure~\ref{fig:hist_knn}, kNN pairs consistently exhibit higher similarity than random pairs in both image and text modalities. This demonstrates that proximity in the generated parameter space correlates with semantic proximity in both vision and language domains.

Together, the global clustering and local semantics alignment analyses provide complementary evidence that PG induces meaningful geometry in parameter space.
These results suggest that PG does not merely produce noisy or arbitrary LoRA parameters. 
Rather than collapsing into a single static adapter (\eg, Shared LoRA) or concentrating into a narrow domain-specific subspace (\eg, Single LoRA), PG organizes instance-conditioned parameter points into a structured conditional parameter family, where local neighborhoods correspond to semantically coherent editing behaviors.
Importantly, this structure emerges without explicit supervision in parameter space, indicating that semantic coherence arises implicitly from instance-conditioned adaptation according to underlying editing semantics.

\subsection{Parameter-Space Geometry of Adaptation}
\label{sec:train_vs_pg}

\begin{wrapfigure}{r}{0.6\textwidth}
    \centering
    \vspace{-20pt} %
    \includegraphics[width=0.58\textwidth]{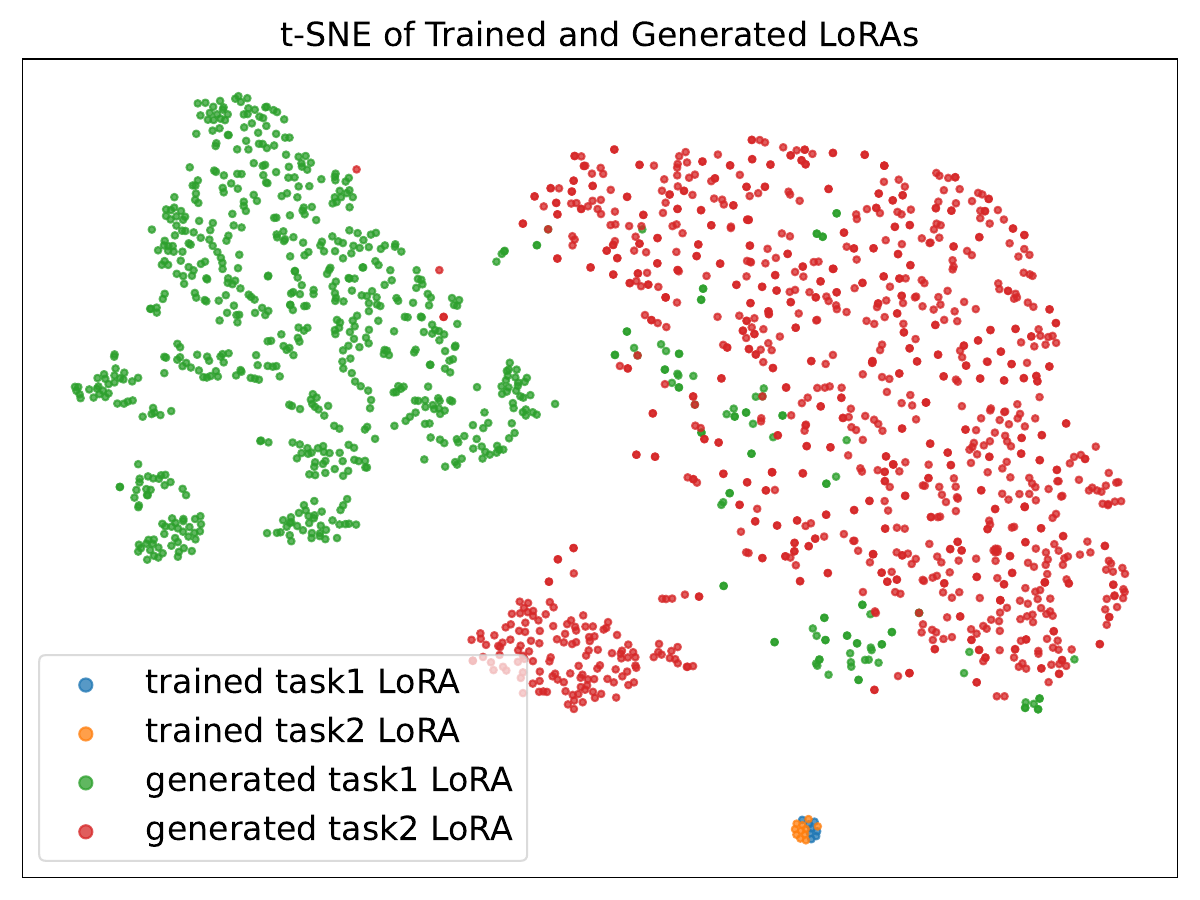}
    \vspace{-8pt}
    \caption{Demonstration of LoRA weights obtained via direct optimization and generated by PG.}
    \vspace{-4pt}
    \label{fig:train_vs_pg_tsne}
\end{wrapfigure}

We compare LoRA weights obtained via direct optimization (SGD) and those generated by PG for two editing tasks.
Figure~\ref{fig:train_vs_pg_tsne} visualizes their distributions using t-SNE.

Directly optimized LoRA weights (blue and orange points) from both tasks collapse into a small region in parameter space, remaining close to their initializations and largely overlapping across tasks.
This suggests that standard training dynamics would confine solutions to a narrow subspace.

In contrast, PG-generated weights occupy a substantially broader and structured region.
Different tasks form clearly separated clusters, while intra-task samples exhibit internal geometric organization.
This indicates that PG learns a conditional parameter manifold rather than converging to a small and random region.
These results suggest that direct optimization under static objectives may confine solutions to a narrow region of parameter space, whereas conditional generation explores a broader structured manifold.
End-to-end conditional generation instead enables exploration of a richer solution family aligned with task semantics.

\section{Related Work}
\label{sec:related_work}

We organize related work around the \emph{adaptation and memory interface} of foundation models, which is the central axis of our framing in Section~\ref{section:intro}. Concretely, we review (i) memory interfaces and routing in foundation models and agentic systems, (ii) continual learning and personalization as stability--plasticity problems under heterogeneous objectives, (iii) parameter-efficient adaptation as \emph{static parameter memory} and its compositional variants, (iv) hypernetworks and conditional parameter generation as operator synthesis primitives, and (v) weight-space modeling and geometry as evidence that parameter updates occupy structured regions. 

The task and modality-specific literature for image editing and personalization was reviewed in Section~\ref{sec:instantiation}, and here we present the more generic picture.

\subsection{Memory in Foundation Models and Agents}
\label{sec:rw_memory_interfaces}

\paragraph{Forms of memory in foundation models.}
Memory in foundation models is distributed across multiple substrates with different persistence, bandwidth, and update costs\citep{huang2026rethinkingmemorymechanismsfoundation}. At the most persistent end, \emph{parameter memory} stores knowledge and behaviors directly in the weights through pretraining or fine-tuning, enabling fast inference-time reuse but making updates expensive and potentially interference-prone when objectives evolve.
Parameter-efficient fine-tuning (PEFT) methods, such as LoRA~\citep{hu2022lora}, reduce update cost by restricting changes to structured low-rank subspaces, but they typically still produce a fixed update applied uniformly at inference.

At the opposite end, \emph{activation-level memory} stores transient information in hidden states, attention contexts\citep{vaswani2017attention}, or cached key--value tensors (in autoregressive systems)\citep{Radford2018ImprovingLU}, enabling fast ``working memory'' behaviors but remaining volatile and bounded by compute and context length.

Between these extremes, \emph{retrieval memory} externalizes information into non-parametric stores (e.g., vector databases or retrieval-augmented generation systems\citep{lewis2021retrievalaugmentedgenerationknowledgeintensivenlp}) that can be queried at inference time, enabling freshness and scalability without modifying $\theta$.
Hybrid systems combine these mechanisms, for example using retrieval to provide evidence and parameter-efficient modules to specialize behavior\citep{guu2020realmretrievalaugmentedlanguagemodel}.
This stratification highlights a core design question: \emph{what should be written into parameters, what should remain in activations, and what should be stored externally and retrieved on demand}.

\paragraph{Structured memory and routing in agents.}
Agentic systems make the memory-interface question explicit because they must decide \emph{what to store}, \emph{where to store it}, and \emph{how to retrieve or apply it} over long horizons.
A common view is that agent memory is naturally multi-typed: episodic traces (interaction histories), semantic abstractions (facts and summaries), procedural knowledge (skills and rules), and resource memories (tools, files, documents)\citep{wang2025mirixmultiagentmemoryllmbased}.
Early agent memory designs often relied on relatively flat stores with uniform retrieval policies, which can suffer from noisy recall and poor controllability as memory grows.
More recent systems emphasize \emph{structured memory components} with explicit \emph{routing, consolidation, and access control} to improve reliability and personalization at scale (e.g., modular memory taxonomies and multi-component routing)~\citep{wang2025selfevolvingagents}.
While these works primarily treat memory as an external store accessed through retrieval and tools, they sharpen a general lesson that we adopt: \emph{memory must be structured and routed, not merely accumulated}.
Our focus differs in \emph{memory substrate}: we consider routing not only over retrieved items, but over \emph{operators} via parameter generation.

\paragraph{Limits of activation-only memory for transformation rules.}
Retrieval and activation-level memory are highly effective when the missing ingredient is \emph{information} that can be expressed as additional context (facts, examples, user history).
However, these interfaces preserve a fixed operator $f(\cdot;\theta)$: they modify what the model sees, not the transformation it implements.
When objectives differ primarily in \emph{transformation rules} (procedural behaviors) rather than in missing context, a fixed operator must internally reconcile conflicting mappings, which can reintroduce compromise effects similar to those in static parameter memory.
This motivates operator-level mechanisms in which the system can express different behaviors as different operators rather than as different prompts or retrieved contexts.

Part~I of HY-WU focuses on this operator-level view and does not benchmark retrieval-based editing baselines; we treat retrieval memory as complementary and defer integrated retrieval--functional-memory evaluations to later parts of the HY-WU series.

\subsection{Continual Learning and Personalization}
\label{sec:rw_continual_learning}

\paragraph{Classical continual learning: stability--plasticity trade-offs.}
Continual learning (CL) studies how to incorporate sequentially arriving tasks without catastrophic forgetting.
A central challenge is the stability--plasticity dilemma: models must remain stable to preserve past knowledge while remaining plastic enough to learn new objectives.
Regularization-based approaches (\eg, EWC~\citep{kirkpatrick2017ewc}, MAS~\citep{aljundi2018memory}) constrain parameter drift on parameters deemed important for prior tasks.
Replay and rehearsal methods~\citep{rebuffi2017icarl,rolnick2019experience,shin2017continual} retain subsets of past data or generate pseudo-samples to approximate previous distributions.
Parameter-isolation and expansion strategies~\citep{rusu2016progressive,yoon2017lifelong} allocate separate modules or subspaces for different tasks to reduce interference.
Across these families, interference arises when heterogeneous objectives induce incompatible gradients on shared parameters, pushing learning toward compromise solutions, as analyzed in Figure~\ref{fig:diagram}.

\paragraph{Continual learning in foundation models.}
With large foundation models, CL has increasingly been implemented through parameter-efficient modules (adapters, LoRA) that keep the backbone largely fixed while accumulating task- or domain-specific updates~\citep{houlsby2019parameter,hu2022lora,pfeiffer2021adapterfusion}.
A common strategy assigns one adapter per task or domain, which reduces destructive interference but leads to growth in stored modules and fragmentation of knowledge, with limited cross-task transfer unless explicit fusion mechanisms are introduced.
Alternatively, approaches that update a shared adapter incrementally reduce storage but reintroduce compromise under heterogeneous objectives.
Related works on model merging and task arithmetic explore integrating multiple task-specific updates through weight-space operations such as averaging, interpolation, or arithmetic~\citep{ilharco2022editing,wortsman2022model}.
These methods provide evidence that update space can be structured and partially compositional, but they typically operate at discrete task granularity and do not directly address \emph{instance-conditioned} routing under continuous heterogeneity.

\paragraph{Personalization as continual adaptation with user-conditional objectives.}
Personalization can be viewed as a fast-timescale CL regime in which objectives drift across users, sessions, or contexts.
A common pattern is to attach lightweight personalization modules (e.g., LoRA-style adapters or hypernetwork-generated adapters) that specialize behavior without retraining the full backbone~\citep{hu2022lora,mahabadi2021hyperformer,ruiz2023dreambooth,ruiz2024hyperdreambooth}.
This paradigm inherits the same structural trade-offs: either maintain separate modules per user (increasing storage and management complexity) or update a shared module (risking interference across heterogeneous user intents).

From our perspective, the core difficulty of both CL and personalization is not only preventing forgetting but avoiding the need to reconcile incompatible objectives within a single shared parameter point.
HY-WU targets this bottleneck by treating adaptation as conditional operator routing rather than repeated overwriting.

\subsection{Parameter-Efficient Adaptation as Static Parameter Memory}
\label{sec:rw_peft}

\paragraph{PEFT methods and shared static updates.}
PEFT methods have become standard mechanisms for specializing large pretrained models under tight compute budgets.
Adapters insert small bottleneck modules into a frozen backbone~\citep{houlsby2019parameter}, prompt and prefix tuning learn continuous conditioning parameters without updating backbone weights~\citep{li2021prefix,lester2021power}, and LoRA injects low-rank updates into selected weight matrices~\citep{hu2022lora}.
Despite architectural differences, these approaches converge to a \emph{single static update}: once training completes, the same learned update is applied to all inference inputs.

\paragraph{Compositional PEFT: fusion, merging, and arithmetic.}
A line of work explores how multiple task-specific updates can be combined without retraining the backbone.
AdapterFusion~\citep{pfeiffer2021adapterfusion} combines multiple trained adapters via a learned fusion module to enable non-destructive multi-task composition.
More broadly, model soups~\citep{wortsman2022model} and task arithmetic~\citep{ilharco2022editing} study composition through averaging and vector operations in weight space, revealing that update spaces often exhibit useful structure and partial compositionality.
These methods typically operate with a \emph{finite, pre-collected} set of modules or checkpoints: inference selects, fuses, or averages among stored updates.
In contrast, HY-WU focuses on generating \emph{instance-conditioned} updates on-the-fly, which can be viewed as continuous routing over an infinite update family rather than discrete selection among a fixed set.

\paragraph{Structural limitation: capacity vs.\ conditionality.}
A recurring ambiguity in adaptation research is whether limitations stem from insufficient trainable capacity (\eg, LoRA rank) or from the static nature of the update.
Increasing trainable parameters (up to full SFT) expands expressivity but still yields a single shared operator at inference.
This motivates separating \emph{capacity} from \emph{conditionality}: static updates represent one global solution that may compromise under heterogeneity, whereas conditional mechanisms can allocate different updates to different inputs.
Our controlled alignment ablations (Table~\ref{tab:instance_ada}) support this distinction by showing that the gains of parameter generation collapse when instance--parameter alignment is removed, indicating that routing, not merely parameter count, drives the effect.

\subsection{Hypernetworks and Conditional Parameter Generation}
\label{sec:rw_hypernetworks}

\paragraph{Classical hypernetworks and dynamic parameterization.}
Hypernetworks~\citep{ha2017hypernetworks} introduced the idea of generating the weights of a target network from a conditioning signal, enabling dynamic parameterization and amortized adaptation.
In this paradigm, weight generation functions as a computational primitive: a learnable generator predicts operator parameters conditioned on context.
Related mechanisms appear in input-conditioned operators such as CondConv~\citep{yang2019condconv} and DynamicConv~\citep{chen2020dynamic}, and in amortization-based meta-learning methods that predict task- or instance-adapted parameters from context embeddings~\citep{santoro2016meta,garnelo2018conditional}.

\paragraph{Checkpoint- and module-conditioned adaptation.}
In large-scale foundation models, conditional selection or generation has also been explored as a mechanism for modular transfer and reuse.
Related work studies selecting or composing pre-trained LoRA modules based on task similarity, treating adapters as reusable building blocks (\eg, LoRAHub-style)~\citep{huang2023lorahub}.
In diffusion-based personalization, concept-specific updates can be learned as adapters, and hypernetwork-style variants predict adapter weights conditioned on concept or subject representations~\citep{ruiz2023dreambooth,ruiz2024hyperdreambooth}.
More efforts~\citep{schurholt2021self,schurholt2022hyper,peebles2022learning,wang2024neural,jin2024conditional,li2024tina,soro2024diffusion,wang2025recurrent,liang2025draganddrop} are devoted to learning pre-collected weight checkpoints via reconstruction loss or augmented with downstream task losses.
This line of work also rely on explicit weight supervision or pre-collected module banks, which may raise scalability and maintenance costs as the number of domains or concepts grows.

Despite these primitives, most existing approaches operate in a \emph{checkpoint-conditioned} regime, where the generator is trained to reconstruct or compose pre-collected adapter weights or checkpoints. In contrast, HY-WU adopts an \emph{on-the-fly training paradigm}: the generator is optimized directly through downstream task loss without explicit weight targets. This shifts parameter generation from a checkpoint reconstruction problem to a functional memory interface that synthesizes instance-conditioned operators from end to end.

\subsection{Weight Space as a Generative Modality and Its Geometry}
\label{sec:rw_weight_space}

Recent work suggests that neural network weights could be treated as a first-class data modality, where collections of trained checkpoints form corpora amenable to representation learning and generative modeling~\citep{wang2025recurrent,liang2025draganddrop}.
HY-WU fits this regime by generating low-rank updates as structured operator modifications.

\paragraph{Geometry that enables learnability and composition.}
Weight-space modeling is feasible in part because learned solutions occupy structured regions rather than arbitrary points.
Empirical findings such as mode connectivity suggest independently trained solutions can be connected by low-loss curves~\citep{garipov2018loss}, and intrinsic-dimension analyses indicate that many objectives admit solutions in surprisingly low-dimensional subspaces~\citep{li2018measuring}.
At the same time, weight space exhibits non-trivial symmetries (e.g., neuron or channel permutations) that complicate naive distance metrics, motivating symmetry-aware alignment methods such as re-basin and weight matching~\citep{ainsworth2022git}.
Together, these properties support the view that high-performing models and updates often lie on structured, learnable weight manifolds.

\paragraph{Compositionality, merging, and interpretability in update space.}
Multiple lines of work study compositional structure in parameter updates, including model merging and ``task vectors'' that exhibit approximately linear semantics under addition and negation~\citep{ilharco2022editing}, as well as checkpoint averaging recipes that improve robustness~\citep{wortsman2022model}.
These results suggest that update spaces can contain semantically meaningful directions and that composition can be performed with simple operations when alignment conditions hold.
Our analysis extends this perspective to \emph{instance-generated} adapter updates, probing whether local neighborhoods and global regions in generated-parameter space correlate with semantic attributes of editing behavior (Section~\ref{sec:parameter_structure}).

\paragraph{Gap and positioning: conditional generation as routed memory over weight space.}
Most prior work frames weight generation as compression, fast adaptation, or model synthesis.
Less commonly is it framed as a \emph{memory interface} with routing semantics for continual personalization: an interface that reduces reliance on overwriting shared weights by allocating different operator updates to different conditions.
HY-WU occupies this gap by treating conditional adapter generation as functional memory, namely routed operator synthesis over a structured update family.
This framing aligns naturally with our conflict-controlled evaluations and alignment ablations, which emphasize that the key difficulty under heterogeneity is not merely storing multiple skills, but avoiding compromise when incompatible objectives would otherwise share a single operator.

\section{Open Roadmap: The HY-WU Series and a Memory-First Agenda}
\label{sec:roadmap}

This report is \textbf{HY-WU Part I}: it establishes functional (operator-level) memory via on-the-fly conditional LoRA generation, and validates the mechanism in a procedural stress test, text-guided image editing.
The broader claim of HY-WU is about \emph{memory interfaces} for foundation models: how new behaviors should be stored, routed, and applied in continual deployment.

To keep the series scientifically grounded, we explicitly separate what Part I demonstrates from what remains open.
Part I shows that \emph{instance-conditioned operator routing} can resolve heterogeneous transformation objectives without collapsing them into a single static compromise.
It does \emph{not} yet benchmark retrieval-based memory baselines for editing, nor does it evaluate long-horizon online continual learning protocols.
This section outlines our planned next steps to complete the memory-first agenda.

\subsection{R1. Retrieval Memory vs.\ Functional Memory: Complementarity, Not Replacement}
\label{sec:roadmap_retrieval_vs_functional}

A key conceptual distinction in Section~\ref{sec:memory_interfaces} is that retrieval memory injects \emph{content} while preserving a fixed operator, whereas functional memory synthesizes the \emph{operator} itself.
Part I focuses on the operator side because image editing is dominated by transformation rules.
A critical next step is to empirically establish when retrieval memory is sufficient, when it is not, and when the two are complementary.

\paragraph{Benchmark triad.}
We propose a minimal but decisive comparison suite across multiple tasks:
\begin{itemize}
    \item \textbf{Retrieval-only (context memory):} retrieve exemplars, user history, or reference images and inject them as additional conditioning without modifying weights.
    \item \textbf{Retrieval + static parameter memory:} retrieval combined with a fixed LoRA/SFT update to test whether additional content mitigates single-point compromise.
    \item \textbf{Retrieval + functional memory (HY-WU):} retrieval supplies content, while HY-WU supplies operator shifts, testing explicit complementarity.
\end{itemize}

\paragraph{Task design principle: facts vs.\ rules.}
To avoid contrived baselines, we recommend splitting evaluations by what the task fundamentally requires:
(i) \emph{facts or exemplars} (where retrieval should dominate), versus
(ii) \emph{transformation rules and procedural control} (where functional memory should dominate).
For generative vision, this naturally maps to: identity and reference consistency (retrieval-heavy) versus rule-driven editing and structural transformations (operator-heavy).
The expected outcome is not that functional memory ``beats'' retrieval memory everywhere, but that it provides the missing axis when the operator must change.

\subsection{R2. From Conditional Routing to Continual Learning: Online Functional Memory Protocols}
\label{sec:roadmap_continual}

The long-term promise of ``instant parameter as functional memory'' is that new behaviors correspond to new regions on a conditional parameter manifold rather than irreversible overwrites of a shared parameter point.
Part I supports this \emph{mechanistically} via conflict-controlled studies and alignment ablations, but it does not yet evaluate true online continual learning (CL).
A next step is to define an online protocol where new objectives arrive sequentially, and the system must improve on them while preserving old behaviors.

\paragraph{Online update targets.}
A practical design axis is \emph{what changes over time}:
\begin{itemize}
    \item \textbf{Generator-only updates:} freeze pretrained backbone $\theta$ and update generator $g_{\phi}$ online.
    \item \textbf{Scoped generators:} maintain user- or domain-scoped memory heads with shared backbone.
    \item \textbf{Hybrid consolidation:} periodically distill frequently used behaviors from functional memory into the backbone, while preserving long-tail behaviors in the generator.
\end{itemize}

\paragraph{Forgetting and consolidation in functional memory.}
Even if $\theta$ is frozen, $g_{\phi}$ can still forget.
We therefore view continual learning for HY-WU as two coupled problems: (i) preserving previously learned \emph{routing and operator synthesis} in $g_{\phi}$, and (ii) preventing the update family from collapsing toward a low-diversity compromise.
Replay, regularization, and modular expansion strategies can be revisited in this generator-centric setting, but evaluated with operator-level diagnostics.

\subsection{R3. Neural Memory Scaling: Allocating Capacity to Functional Memory}
\label{sec:roadmap_scaling}

A central hypothesis behind ``memory-first'' design is that scaling should not only mean increasing backbone size.
Instead, we make a bold conjecture: \textbf{jointly scaling a backbone together with a memory module can be more compute- and data-efficient than scaling a monolithic backbone alone.}

The intuition is structural rather than cosmetic. Monolithic scaling must amortize long-tail and conflicting objectives into a single parameter point, which encourages compromise and interference; functional memory allocates \emph{conditional} operator capacity, so rare or conflicting behaviors need not be baked into a shared weight point.
Under this view, backbone parameters should primarily encode stable invariants, while functional memory supplies instance-conditioned operator shifts, yielding a second scaling axis that targets heterogeneity directly rather than hoping it averages out.

\paragraph{A concrete scaling study.} Part I introduces architectural tools (tokenization and factorized attention) that make such memory scalable, but does not yet present scaling curves. We propose an explicit scaling grid holding the backbone fixed while scaling functional memory along multiple axes:
\begin{itemize}
    \item Generator capacity: depth/width of the Neural Network Transformer.
    \item Update bandwidth: number of adapted modules and LoRA rank $r$.
    \item Token budget: $(l,s,r,d)$ induced by tokenization choices.
    \item \emph{Conditional diversity}: an empirical proxy for the ``conditional entropy'' of generated updates, measured via covariance spectra or intrinsic dimension of $\Delta\theta(x)$ over deployment samples.
\end{itemize}
The goal is to quantify when additional backbone parameters saturate, but additional functional memory capacity continues to improve controllability, conflict robustness, and personalization.

\paragraph{Evaluation beyond ``average benchmark performance".}
For generative systems, scaling should be evaluated not only by mean quality but also by \emph{behavioral coverage}: the ability to execute diverse, highly individualized transformation rules without compromise.
This naturally pairs with the manifold analyses in Section~\ref{sec:parameter_structure} and \ref{sec:train_vs_pg}, turning them into quantitative scaling diagnostics.

\subsection{R4. Beyond LoRA: General Operator Interfaces for Functional Memory}
\label{sec:roadmap_beyond_lora}

LoRA is an effective conduit for Part I because it is structured, parameter-efficient, and easy to inject at inference time.
However, a memory-first framework should not be locked to a single adapter parameterization.
We propose extending functional memory to richer operator interfaces:
\begin{itemize}
    \item Alternative PEFT parameterizations (e.g., adapters, gating, input-conditioned modulation).
    \item Interface specialization per modality (e.g., convolutional deltas for vision backbones, attention-projection deltas for transformers).
    \item Hierarchical operator synthesis, where coarse updates control high-level behavior and fine updates control local edits.
\end{itemize}
This direction also enables sharper theoretical questions: which operator subspaces are most expressive per unit bandwidth, and which yield the best controllability per unit compute.

\subsection{R5. Long-Horizon Multimodal Memory: From Images to Video Personas and Agents}
\label{sec:roadmap_multimodal}

Part~I establishes operator-level functional memory in a controlled, transformation-centric regime.
However, the full implications of a memory-first architecture become most visible in longer-horizon and multimodal settings.
We therefore view the current study as a foundational step and outline the following concrete extensions within the HY-WU series.

\paragraph{R5.a: Video persona and identity-consistent generation.}
Video introduces persistent identity and long-horizon coherence constraints that amplify the limits of static adaptation.
Two distinct memory pressures emerge simultaneously: 
(i) long-term identity consistency across frames and sessions, and 
(ii) temporally stable transformation rules applied to evolving content.
These pressures naturally separate memory substrates: retrieval-style memory is well suited for maintaining persistent identity references and history, while functional memory governs operator-level transformations.
A decisive next study will benchmark retrieval-only, functional-only, and combined retrieval+functional memory under persistent persona constraints, evaluating stability, controllability, and interference across extended sequences.

\paragraph{R5.b: Agentic multimodal systems and procedural skills.}
Agentic systems surface a broader class of procedural memory requirements, including tool-use policies, planning heuristics, preference-conditioned behaviors, and task-specific reasoning strategies.
In such settings, functional memory can synthesize operator shifts conditioned on tool state, environment feedback, or user profile, while retrieval memory supplies factual or episodic context.
This environment provides a principled test bed for functional memory under sequential skill acquisition, where the central question becomes whether routing over a learned update family can reduce destructive interference as new procedural competencies accumulate.

\subsection{R6. Safety, Privacy, Governance, and System Implications}
\label{sec:roadmap_safety}

A memory-first system introduces a new control surface: the generated operator update $\Delta\theta(x)$.
This enables new safety and governance mechanisms, but also new risks.

\paragraph{Auditable and controllable memory.}
We propose designing explicit controls over functional memory, for example:
(i) bounding the norm or rank of $\Delta\theta(x)$,
(ii) adding safety filters on the condition $c(x)$ and on generated updates,
(iii) supporting rollback and ``memory deletion'' by manipulating generator states or routing rules.

\paragraph{Privacy and memory scope.}
If personalization is implemented through functional memory, it becomes crucial to define whether memory is ephemeral (session-level) or persistent (user-level), and how it is stored.
A rigorous deployment plan should include threat modeling for leakage through generated parameters and evaluate privacy-preserving training or encryption-compatible storage for user-scoped memory.

\paragraph{Systems Roadmap: Latency, Caching, and Memory Banks.} Finally, moving from proof-of-concept to deployment requires treating functional memory as a systems interface.
Key directions include:
(i) caching generated adapters for repeated conditions,
(ii) amortizing generation across diffusion timesteps,
(iii) batching and sequence-parallel strategies for long parameter tokens,
and (iv) memory banks of prototypical updates that can be refined by generation.

\clearpage
\appendix
\section{Project Contributors}

\begin{itemize}[leftmargin=*] %

    \item \textbf{Project Sponsor:}  Linus

    \item \textbf{Project and Tech Lead:} Qinglin Lu, Kai Wang

    \item \textbf{Core Contributors:} Mengxuan Wu, Xuanlei Zhao, Ziqiao Wang, Ruicheng Feng, Atlas Wang\footnote{Not affiliated with Tencent. Contributions were made independently as an individual researcher during personal time, focusing on key idea formulation and experimental design, and should not be interpreted as representing any organization.}, Kai Wang

    \item \textbf{Contributors:} Jiale Tao, Junshu Tang, Haoyu Yang,  Zhentao Yu, Xiang Wen, Chunyu Wang, Shuai Shao, Han Hu

    \item \textbf{Data and Benchmarking Assistance:} Jihong Zhang, Qiuyong Xiao, Liyu Wang, Jiaxin Lin, Feng Lu, Lei Wang, Shiyu An, Shuo Jiang, Zhichao Hu

    \item \textbf{Outside Contributors and Advisors:} Ziheng Qin (NUS), Zekai Li (UCSD),  Zhaoyang Zeng (SYSU), Shangchen Zhou (NTU), Xindi Wu (Princeton), Zhaopan Xu (NUS), Shengbin Huang (NTU), Xiaojiang Peng (SZTU), Yuzhang Shang (UCF),  Damian Borth (University of St. Gallen), Huazhe Xu (THU), Michael Bronstein (Oxford), Yang You (NUS)

\end{itemize}

\begin{figure*}[t]
    \centering
    \includegraphics[width=\linewidth]{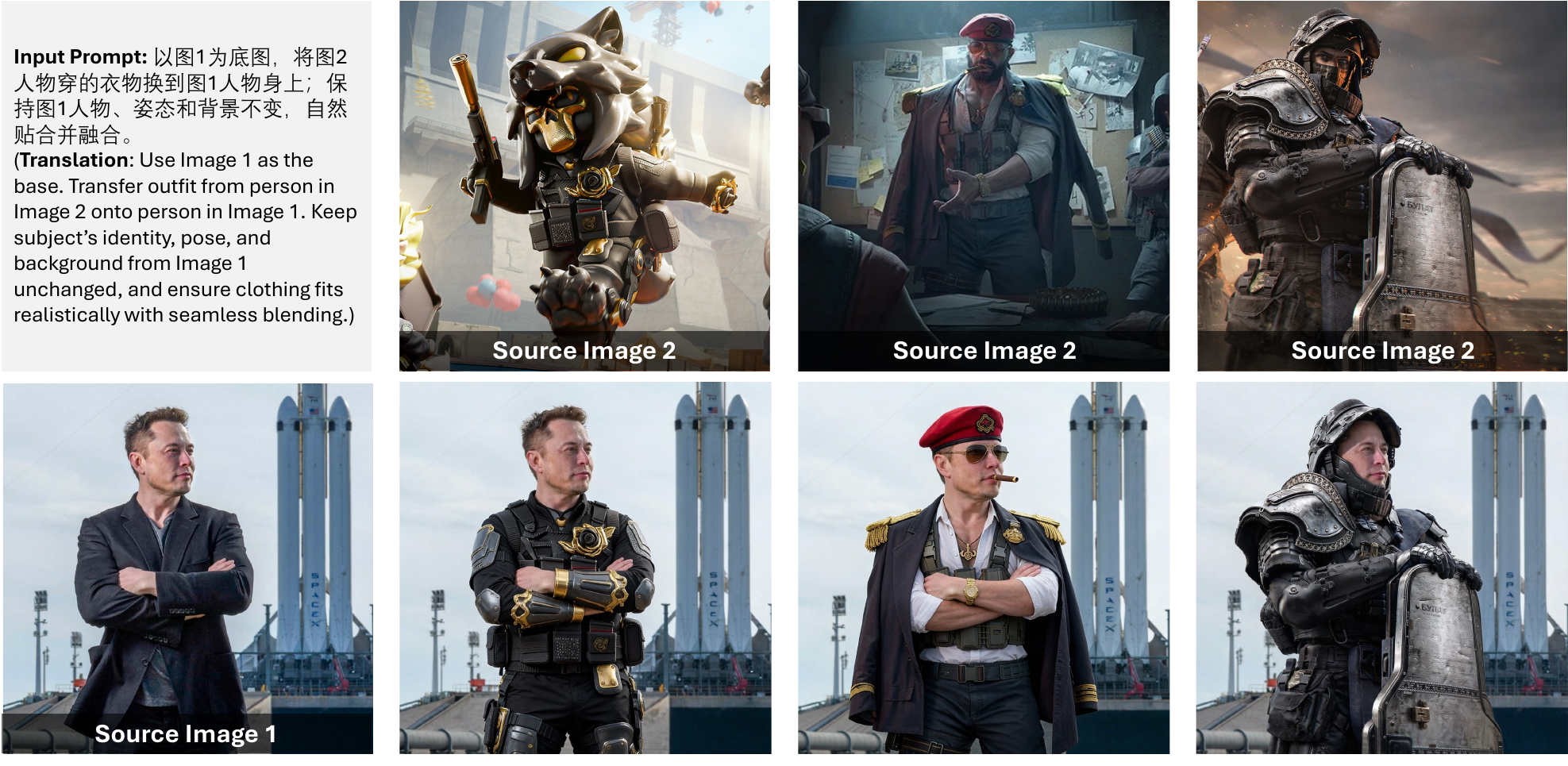}
    \includegraphics[width=\linewidth]{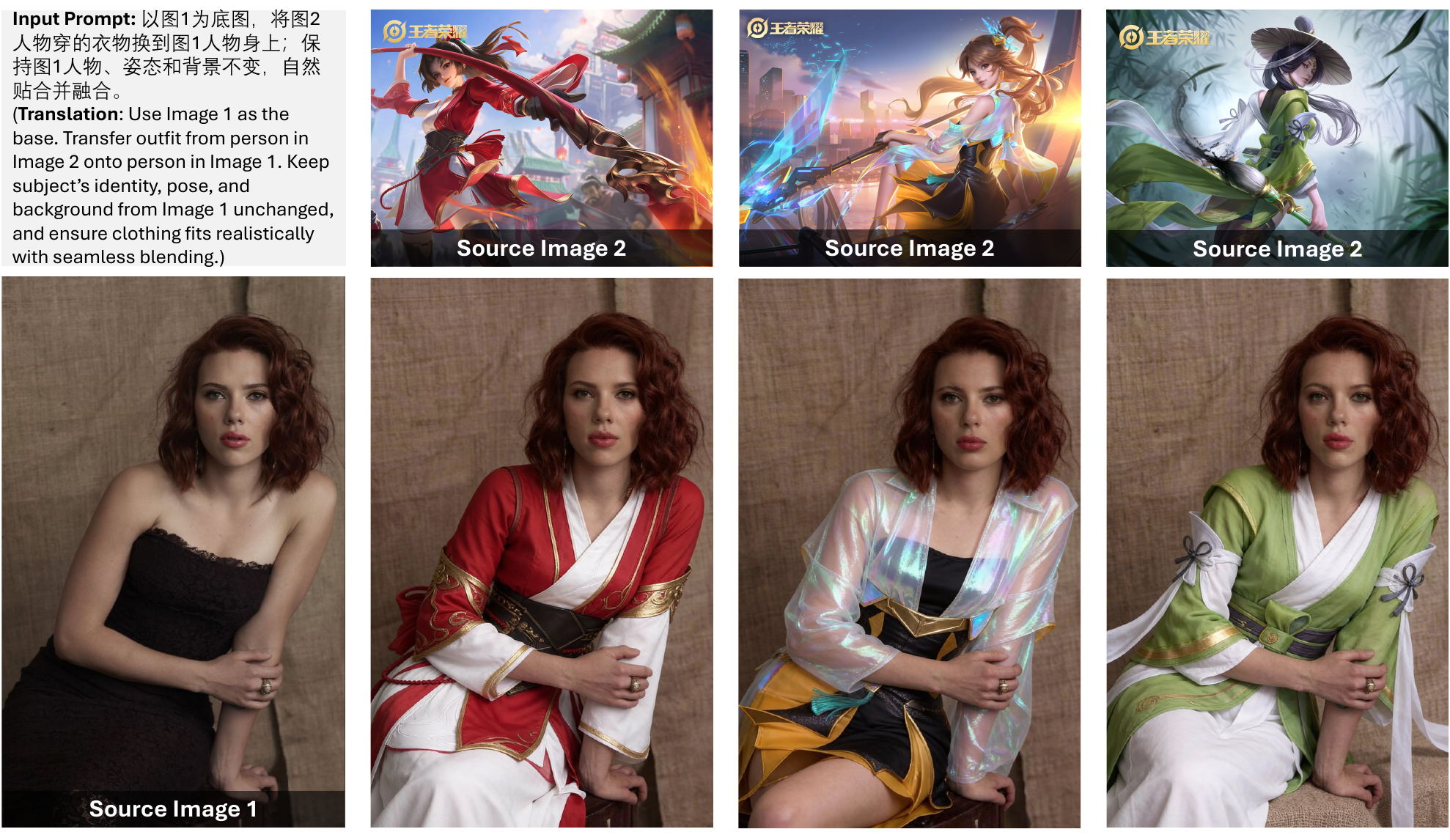}
    \caption{This visualization clearly demonstrates the ability of our HY-WU to preserve key invariants and generate high-quality and creative edited images without identity drift.
}
    \label{fig:showcase_gaming1}
\end{figure*}

\begin{figure*}[t]
    \centering
    \includegraphics[width=\linewidth]{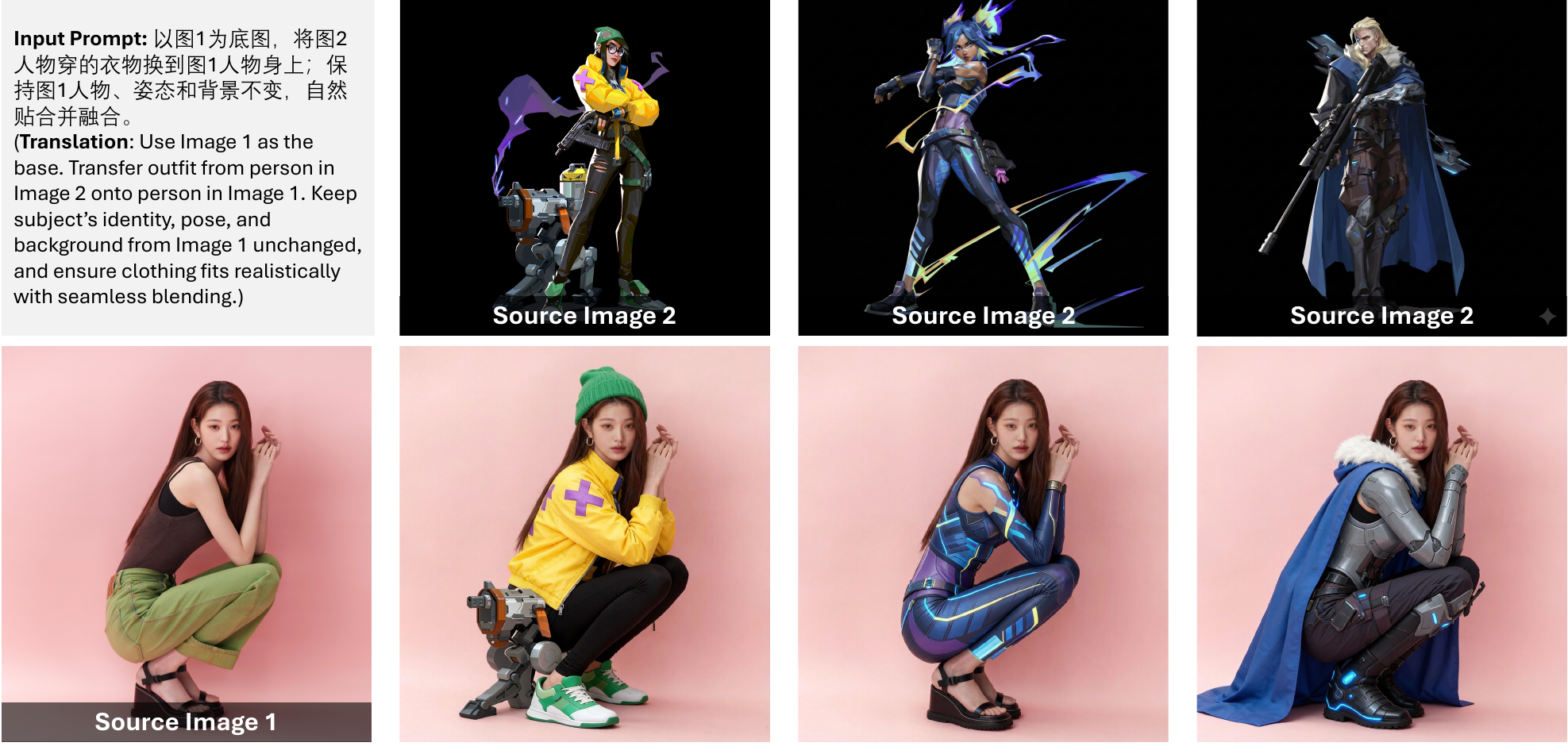}
    \includegraphics[width=\linewidth]{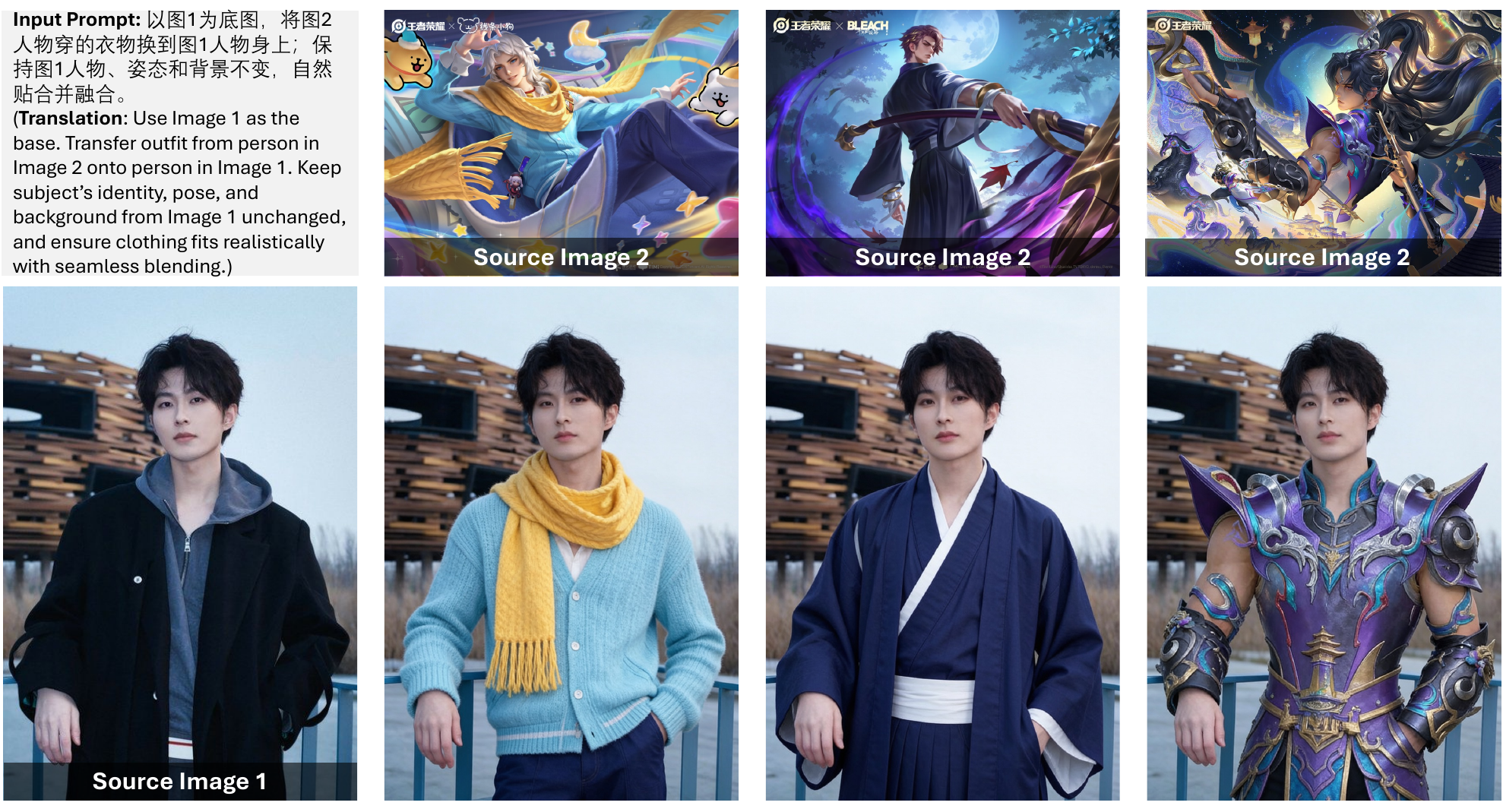}
    \caption{This visualization clearly demonstrates the ability of our HY-WU to preserve key invariants and generate high-quality and creative edited images without identity drift.
}
    \label{fig:showcase_gaming2}
\end{figure*}

\begin{figure*}[t]
    \centering
    \includegraphics[width=\linewidth]{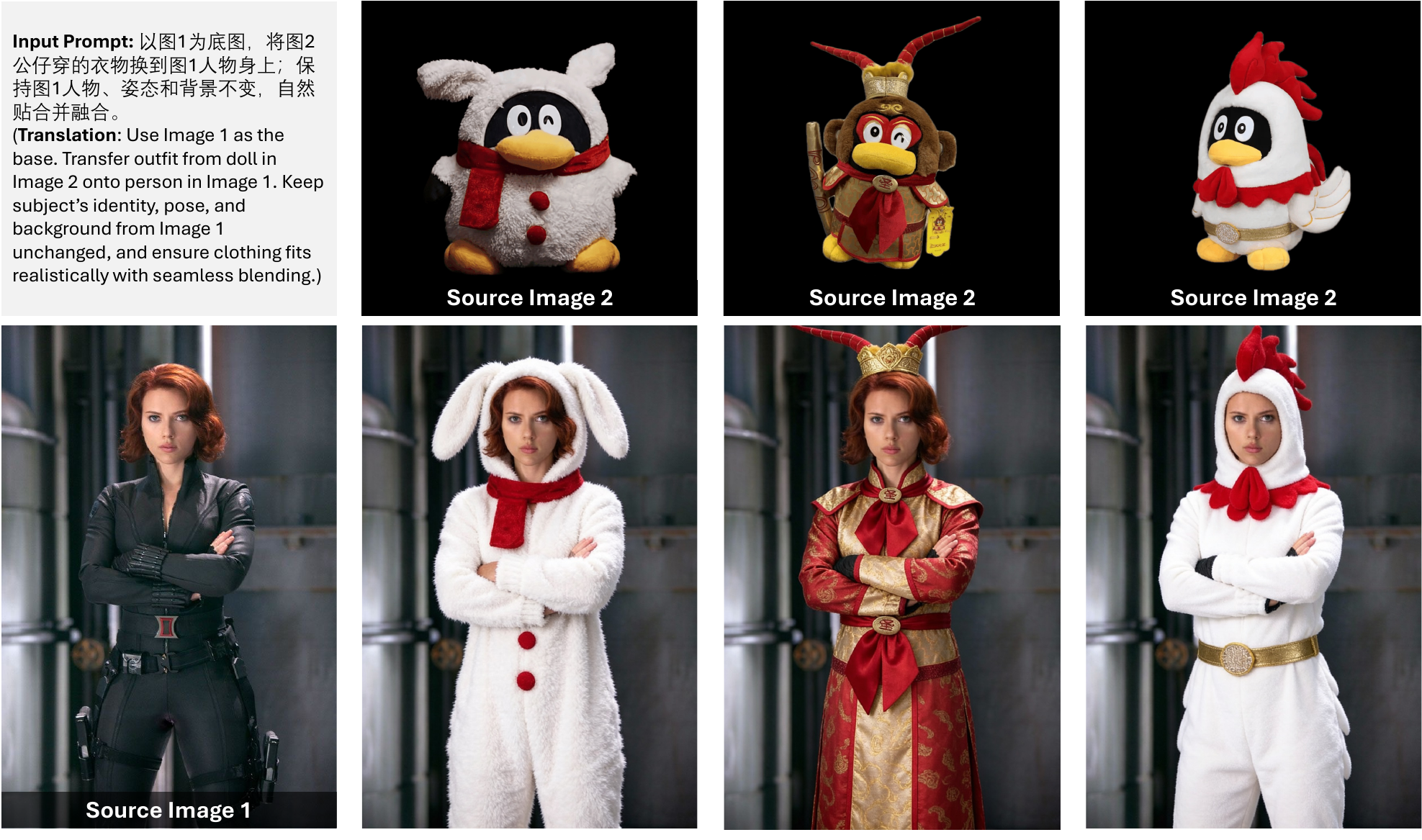}
    \includegraphics[width=\linewidth]{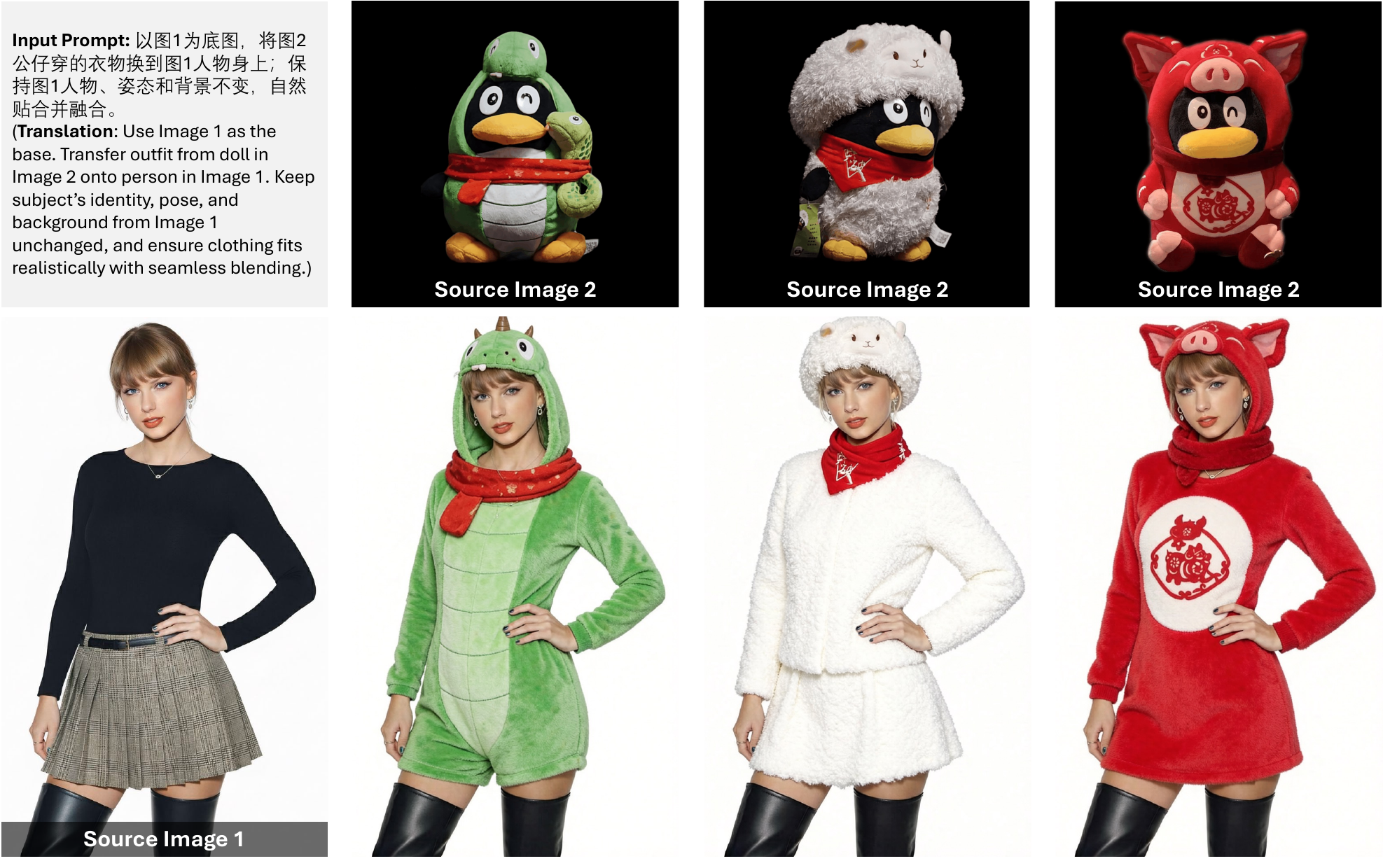}
    \caption{This visualization clearly demonstrates the ability of our HY-WU to preserve key invariants and generate high-quality and creative edited images without identity drift.
}
    \label{fig:showcase_doll}
\end{figure*}

\begin{figure*}[t]
    \centering
    \includegraphics[width=\linewidth]{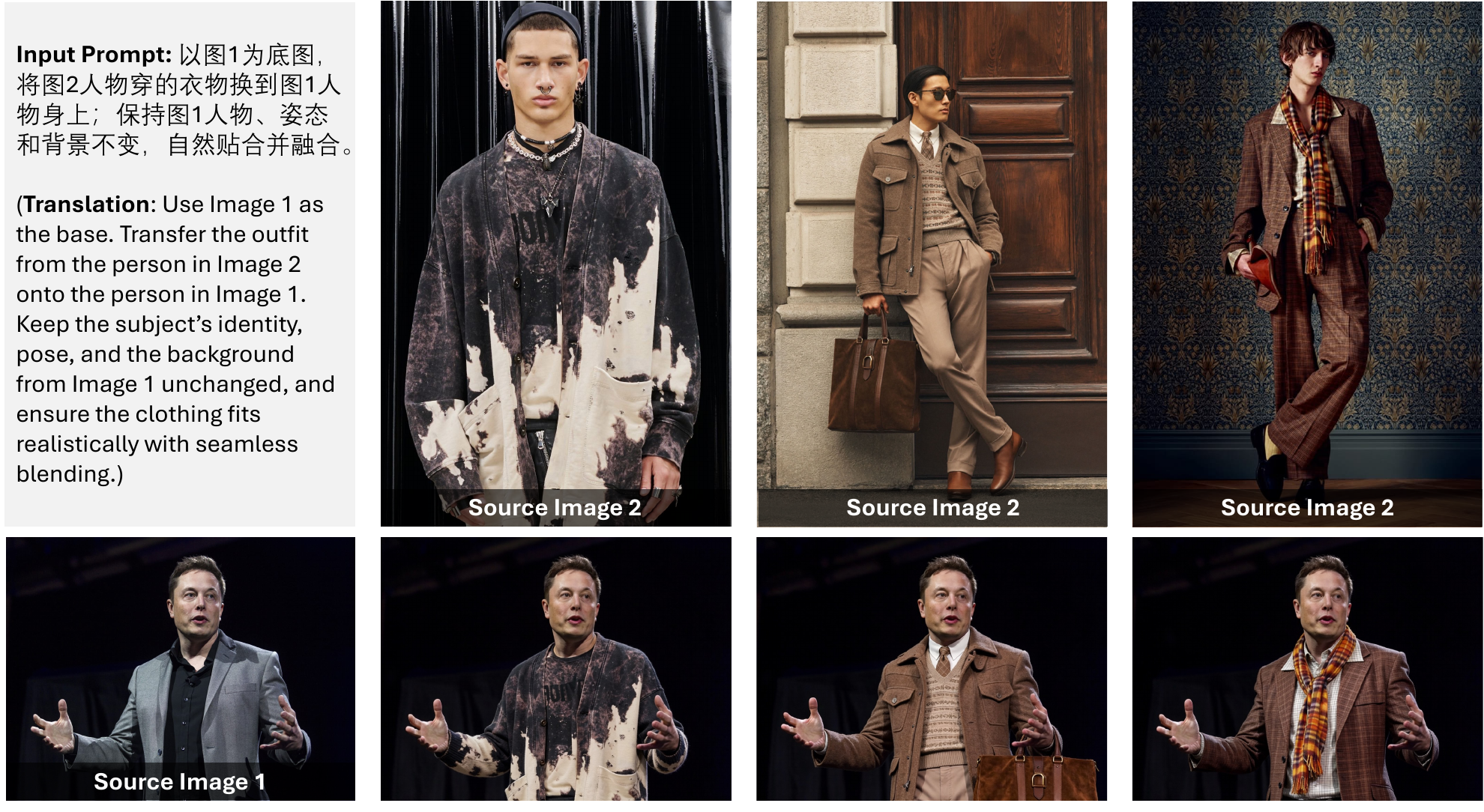}
    \includegraphics[width=\linewidth]{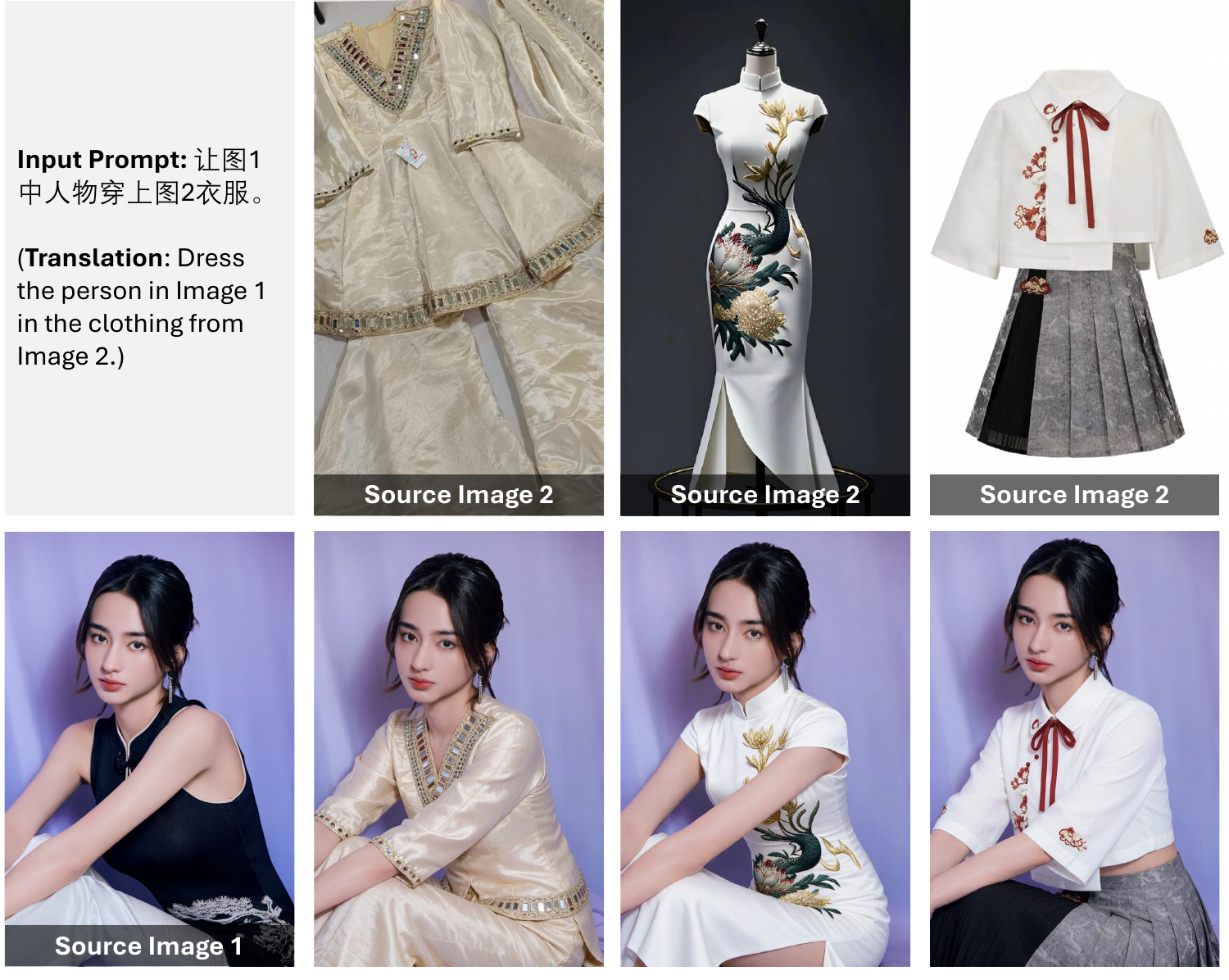}
    \caption{This visualization clearly demonstrates the ability of our HY-WU to preserve key invariants and generate high-quality and creative edited images without identity drift.
}
    \label{fig:showcase_tryon}
\end{figure*}

\begin{figure*}[t]
    \centering
    \includegraphics[width=\linewidth]{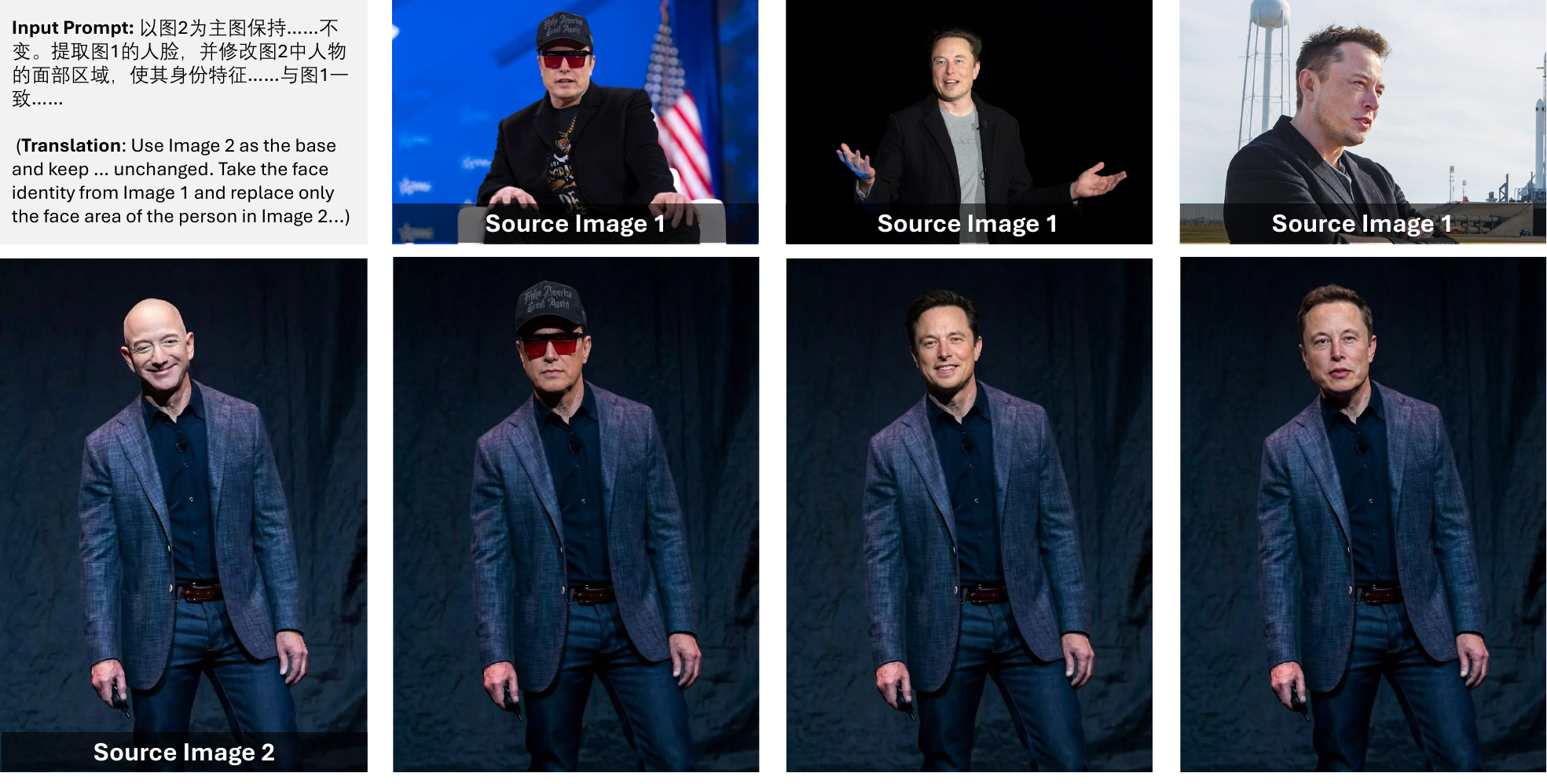}
    \includegraphics[width=\linewidth]{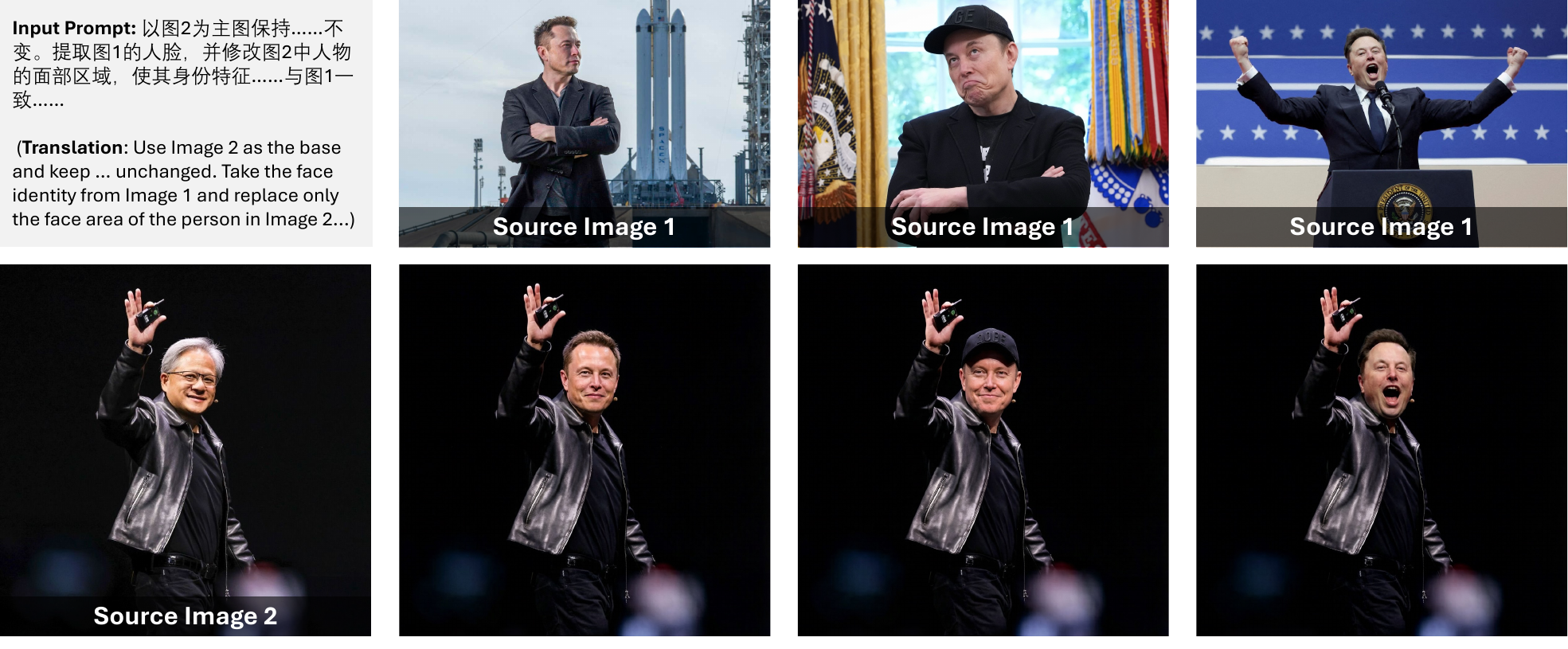}
    \caption{This visualization clearly demonstrates the ability of our HY-WU to preserve key invariants and generate high-quality and creative edited images without identity drift.
}
    \label{fig:showcase_faceswap}
\end{figure*}

\clearpage
\newpage

\label{references}
\addcontentsline{toc}{section}{References}
\bibliographystyle{assets/plainnat}
\bibliography{paper}

\end{document}